\documentclass[final,5p,times,twocolumn]{elsarticle}

\usepackage{algorithm}  
\usepackage{algorithmic}  
\usepackage{bm}
\usepackage{ bbold }
\usepackage{multirow}
\usepackage{caption}
\usepackage{subfigure}
\usepackage{booktabs}
\usepackage{float}
\usepackage{booktabs}

\usepackage{color}
\usepackage{amsmath}
\usepackage{amssymb}
\usepackage{ dsfont }
\usepackage{ upgreek }
\usepackage{booktabs}
\usepackage{multirow}
\usepackage[table,xcdraw]{xcolor}
\usepackage{subfigure}

\begin{document}
	
\begin{frontmatter}
		
		

		\title{Deep Forest with Hashing Screening and Window Screening }
			\author[1]{Pengfei Ma}
		\address[1]{School of Artificial Intelligence, Hebei University of Technology,
			Tianjing,
			300401, 
			China}
		
		\author[1,2]{Youxi Wu \corref{mycorrespondingauthor}}
		
		\cortext[mycorrespondingauthor]{Corresponding author}
		\ead{wuc567@163.com}
		\address[2]{Hebei Key Laboratory of Big Data Computing,
			Tianjing,
			300401, 
			China}
		
		\author[3]{Yan Li}
		\address[3]{School of Economics and Management, Hebei University of Technology,
			Tianjing,
			300401, 
			China}
		
		\author[4]{Lei Guo}
		\address[4]{State Key Laboratory of Reliability and Intelligence of Electrical Equipment, Hebei University of Technology,
			Tianjing,
			300401, 
			China}
		
		\author[5]{He Jiang}
		\address[5]{School of Software, Dalian University of Technology,
			Dalian,
			116023, 
			China}
		
		\author[6]{Xingquan Zhu}
		\address[6]{Department of Computer \& Electrical Engineering and Computer Science, Florida Atlantic University,
			FL,
			33431, 
			USA}
		\author[7,8]{Xindong Wu}
		\address[7]{%
			Key Laboratory of Knowledge Engineering with Big Data (the Ministry of Education of China), Hefei University of Technology,
			Hefei,
			230009,
			China}
		\address[8]{%
			Mininglamp Academy of Sciences, Mininglamp Technology,
			Beijing,
			100084,
			China}

	\begin{abstract}
		As a novel deep learning model, gcForest has been widely used in various applications. However, the current multi-grained scanning of gcForest produces many redundant feature vectors, and this increases the time cost of the model. To screen out redundant feature vectors, we introduce a hashing screening mechanism for multi-grained scanning and propose a model called HW-Forest which adopts two strategies, hashing screening and window screening. HW-Forest employs perceptual hashing algorithm to calculate the similarity between feature vectors in hashing screening strategy, which is used to remove the redundant feature vectors produced by multi-grained scanning and can significantly decrease the time cost and memory consumption. Furthermore, we adopt a self-adaptive instance screening strategy to improve the performance of our approach, called window screening, which can achieve higher accuracy without hyperparameter tuning on different datasets. Our experimental results show that HW-Forest has higher accuracy than other models, and the time cost is also reduced.
	\end{abstract}
	

\end{frontmatter}
	
	%

\section{Introduction}
%
%
%
%

Deep learning \cite {Yann2015} can obtain the better performance than classical machine learning methods \cite{cheng2021ifo,wu2021tkd,liu2016neu,wu2017jcst} and has been widely applied in various fields, such as image processing \cite{zhang2020ins}, speech recognition \cite{speech2018}, and text processing \cite{qiang2020tkde}. However, deep learning models based on neural network involve large numbers of hyperparameters and depend on a parameter tuning process. More importantly, theoretical analyses of neural network models are extremely difficult. There are therefore several reasons for exploring the development of a novel deep learning model. Based on the recognition that layer-by-layer processing and in-model feature transformation are key aspects of neural network models, Zhou and Feng \cite{zhou2019nsr} proposed a novel deep learning model called gcForest.

Inspired by neural network models, gcForest adopts a cascade structure to process data, and contains many random forests at each level of the cascade structure. Multi-grained scanning is employed, in which, as an in-model feature transformation mechanism, a sliding window is used to scan high-dimensional raw data. The main advantage of gcForest is that it does not need backpropagation at the training stage, which makes it much easier to train than the neural network model. Another advantage is that  the complexity of gcForest can be automatically determined and adapted to different training datasets. Recently, gcForest has achieved excellent performance on a broad range of tasks, such as cancer detection \cite{su2019met} and remote sensing \cite{yang2018grs}. However, although existing experimental results show that larger models may provide higher accuracy, gcForest is difficult to implement with large models. The reason for this is that multi-grained scanning transforms a raw instance of high-dimension data into hundreds or thousands of novel image instances, and inputs them into all levels of the cascade structure. Thus, gcForest imposes high time cost and memory consumption to realize learning. 

To reduce the time costs and memory requirements of gcForest, Pang et al. \cite{pang2020TKD} proposed gcForestcs, in which a confidence screening mechanism is introduced into the general framework of gcForest. 
Confidence screening mechanisms select instances according to their confidence, where instance confidence is produced by their prediction vectors. More specifically, all instances are divided into two subsets: high-confidence and low-confidence. Instances in the high-confidence set are used to produce a final prediction at the current level, while the instances in the low-confidence set are passed to the next level. The key aspect of gcForestcs is confidence screening, and its threshold is produced by a simple and effective rule. However, this rule cannot avoid the mispartitioning of instances, in which some low-accuracy instances are classified into the high-confidence set. To tackle this issue, we previously proposed DBC-Forest \cite{ma2021neu}, in which binning confidence screening is used to replace confidence screening. More concretely, DBC-Forest packs all instances into bins based on their confidence values, and adopts the instance accuracy of the bins to avoid mispartitioning of instances. Experimental results showed that DBC-Forest achieved better performance than gcForestcs.

However, current deep forest models have two issues. First, multi-grained scanning can produce some redundant feature vectors which increase time cost. The reason for producing them is that some instances have the same features. Second, in cascade structure, the thresholds of these confidence screening mechanisms depend on tuning hyperparameters. In other words, they cannot achieve the best performance when the hyperparameters are not suitable for some datasets. To improve the performance of the deep forest models, we propose HW-Forest, which adopts two screening mechanisms: hashing screening and window screening. Hashing screening reduces the number of redundant feature vectors based on their perceptual hashing values, and can also be applied to other deep forest models, since the mechanism is an improvement on gcForest. Furthermore, inspired by the binning confidence screening process used in DBC-Forest, a self-adaptive mechanism called window screening is proposed, in which a window is adopted to screen instances. Unlike DBC-Forest, window screening produces thresholds automatically for different datasets without requiring hyperparameter tuning. 

The contributions of this paper can be summarized as follows.

\begin{enumerate}
	\item To improve the performance and efficiency of the deep forest algorithm, we propose a novel deep forest model called HW-Forest which uses two screening mechanisms: hashing screening and window screening.
	
	\item In HW-Forest, hashing screening is used to remove the redundant feature vectors produced by multi-grained scanning, which significantly decreases the time cost and memory consumption.
	
	\item A self-adaptive confidence screening is used to realize window screening, which can achieve higher accuracy without hyperparameter tuning on different datasets.
	
	\item Our experimental results show that the proposed hashing screening mechanism can not only reduce the running time of HW-Forest, but also reduce the running time of other deep forest methods. More importantly, HW-Forest yields better performance than other state-of-the-art approaches.
	
\end{enumerate}

The remainder of the paper is organized as follows. Section 2 introduces related work on the deep forest algorithm. Section 3 describes perceptual hashing and DBC-Forest. Section 4 introduces HW-Forest and its two screening mechanisms. Section 5 presents experimental results. Finally, conclusions are drawn in Section 6.

\section{RELATED WORK}

As a novel deep learning model, deep forest gives higher accuracy than deep neural networks, and it has therefore been applied in many fields, such as medicine, electricity, emotion recognition, agriculture, topic classification for tweets, image retrieval, and economics \cite{wu2021kbs}. In medicine, Sun et al. \cite{sun2020bhi} proposed AFS-DF, in which a feature selection mechanism was adopted to classify cases of COVID-19. To improve the accuracy of classification of cancer subtypes, Guo et al. \cite{guo2018bmc} and Dong et al. \cite{dong2019aps} proposed the boosting cascade deep forest and multi-weighted gcForest algorithms, respectively. In electricity. Wang et al. \cite{wang2021enb} proposed a novel voltage-current trajectory classification model based on deep forest. In emotion recognition, Cheng et al. \cite{cheng2021jbh} applied deep forest to emotion recognition based on electroencephalography. Fang et al.\cite{fang2020fne} proposed a multi-feature deep forest approach in which a multi-feature structure was employed to extract power spectral density and differential entropy features. In agriculture, Zhang et al. \cite{zhang2020sap} combined hyperspectral imaging technology with deep forest to identify frost-damaged rice seeds. In tweet topic classification, a semantic deep forest technique was developed \cite{dao2021ins} in which a semantic layer was introduced into the cascade structure of deep forest. In image retrieval, Zhou et al. \cite{zhou2019par} proposed a deep forest hashing approach to learn shorter binary codes. In economics, Ma et al. \cite{ma2020par} proposed a cost-sensitive deep forest algorithm that assigned a specific cost for each misclassification, to improve the accuracy of deep forest in the area of price prediction.

However, the performance of deep forest needs to be further improved, since it is inefficient on datasets with larger numbers of instances. The most representative of the improved deep forest models is gcForestcs \cite{pang2020TKD}, in which confidence screening was adopted to improve the efficiency. Inspired by gcForestcs, many other approaches have been proposed, such as DBC-Forest \cite{ma2021neu}, AWDF \cite{utk2020ted}, and, MLDF \cite{yang2020cai}. In the binning confidence screening deep forest algorithm \cite{ma2021neu}, a binning confidence mechanism was adopted to improve the accuracy of gcForestcs. In an approach called adaptive weighted deep forest \cite{utk2020ted}, a special weighted scheme was utilized that could adaptively weight each training instance at each cascade level. The multi-label deep forest algorithm \cite{yang2020cai} employed a measure-aware feature reuse mechanism to retain valuable presentation in the previous layer, and a measure-aware layer growth was adopted to limit the complexity of the model with different performance measures.

In the models described above, less attention is paid to the redundant feature vectors, which not only increase the time cost but also decrease the accuracy of the model. To further improve the performance of deep forest, we propose a novel model called HW-Forest, in which hashing screening and window screening are applied. Our experimental results show that HW-Forest achieves better performance than other state-of-the-art models.

\section{Preliminaries}

Section 3.1 introduces perceptual hashing algorithms. Section 3.2 reviews the principles underlying DBC-Forest, including the use of binning confidence screening.
\subsection{Perceptual Hashing}

Perceptual hashing \cite{zhang2021tkd} is an effective approach for image retrieval in which images are converted into binary sequences for retrieval. The most representative perceptual hashing algorithms are aHash \cite{yang2021tnn}, pHash \cite{xu2021tmu}, and dHash \cite{fang2018neu}. To process data efficiently, we employ aHash to screen out redundant feature vectors, using a hashing screening mechanism. The main steps of aHash are as follows:

Step 1: Resize the image (this is a fast way to match an image of any size). 

Step 2: Convert the resized image to grayscale.

Step 3: Compute the mean value of the grayed image pixels.

Step 4: Compare each pixel value with the mean value, and set any pixels larger than the mean to one, and the other pixels to zero.

Step 5: Combine the binarization results into binary sequences.

\begin{figure}
	\centering
	\includegraphics[width=0.8\linewidth]{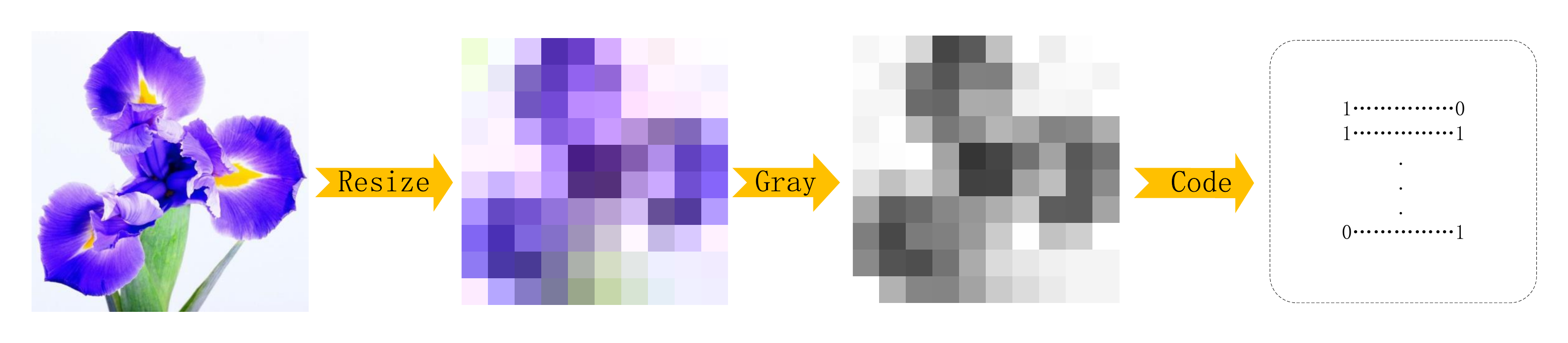}
	\caption{Processing of the image with aHash}
	\label{fig:figure1}
\end{figure}

Finally, aHash retrieves images by comparing their binary sequences. The process used by the aHash algorithm is illustrated in Figure  \ref{fig:figure1}.

\subsection{DBC-Forest}

As an improved version of gcForestcs, DBC-Forest adopts a binning confidence screening mechanism to avoid the mispartitioning of instances with low accuracy into high-confidence regions. More specifically, the binning confidence screening mechanism packs all instances into bins, and DBC-Forest then calculates the accuracy of all bins to locate mispartitioned instances. This mechanism can be summarized as follows.

Step 1: Calculate the confidence of each instance at the current level. We utilize confidence as the criterion for selecting instances, and this is generated by the estimated class vector. For example, if the estimated class vector for instance \textbf{\emph{x}} is (0.5, 0.3, 0.2), its confidence is 0.5.

Step 2: Rank instances based on their confidences. We then obtain instances $(\bm{x}_{s1},\bm{x}_{s2},\cdots,\bm{x}_{sm})$, where $\bm{x}_{s1}$ is the highest confidence instance, and put these instances into $k$ bins. Each bin $\bm{b}_t$ (1$\le t \le k$ ) can be obtained according to Equation (1):

\begin{equation}
	(\bm{x}_{s(\frac{m(t-1)+1}{k})},\bm{x}_{s(\frac{m(t-1)+2}{k})}, \dots, \bm{x}_{s(\frac{mt}{k})})\in \bm{b}_t 
\end{equation}

Step 3: Calculate the accuracy for each bin. The accuracy $\bm{P}_t$ of $\bm{b}_t$ is calculated according to Equation (2):
\begin{equation}
	\bm{P}_t =\frac{\sum_{i=\frac{m}{k}\times(t-1)+1}^{m\times \frac{t}{k}}\mathds{1}[p(\bm{x}_{si})=\bm{y}_{si}]}{m/k}
\end{equation}
where $p({\bm{x}_{si}})$ is the prediction of $\bm{x}_{si}$ at the current level, and $\bm{y}_{si}$ is the label of $\bm{x}_{si}$.

Step 4: Determine the threshold. In DBC-Forest, the accuracy of each bin is compared with the target accuracy $TA$ from 1 to $k$. If $\bm{P}_{j+1}$ is the first bin for which the accuracy is lower than $TA$, then we utilize the confidence of $\bm{x}_{s(m\times \frac{j}{k})}$ as the threshold.

Step 5: Compare the confidences of the instances with the threshold. If the confidence of the instance is larger than the threshold, then the instance is used to produce a final prediction at the current level; otherwise, the instance is passed to the next level.

\section{Proposed Method}

Section 4.1 introduces the hashing screening strategy, and Section 4.2 presents the window screening strategy. Section 4.3 describes HW-Forest and illustrates its framework.

\subsection{Hashing Screening}

\begin{figure}
	\centering
	\includegraphics[width=0.7\linewidth]{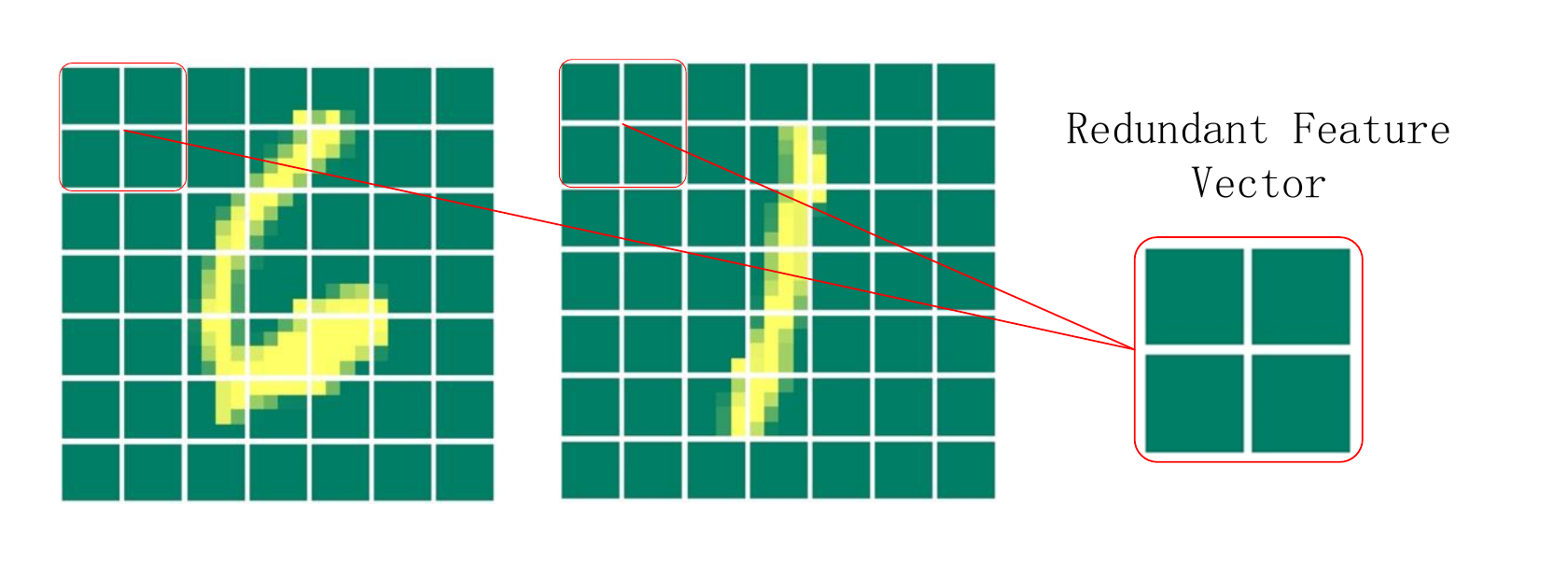}
	\caption{The production of redundant feature vectors. We select two instances from the MNIST dataset, and assume the size of the slide window is 2×2.}
	\label{fig:figure2}
\end{figure}

Multi-grained scanning is a structure for processing high-dimensional data that can achieve excellent performance on image data. However, the time cost of multi-grained scanning is increased by the presence of redundant feature vectors, which are produced by sampling the same features from instances. Figure \ref{fig:figure2} shows the production of a redundant feature vector by a sliding window.

To reduce the number of redundant feature vectors, we apply the aHash algorithm to evaluate the similarity between feature vectors and eliminate the high similarity feature vectors. This process can be summarized as follows.

Step 1: Group feature vectors according to their locations, which gives $(\bm{v}_1, \bm{v}_2,\cdots , \bm{v}_r)$ according to Equation (3):

\begin{equation}
	(\bm{f}_{r,1},\bm{f}_{r,2},\dots,\bm{f}_{r,m})\in\bm{v}_r
\end{equation}

\begin{figure}
	\centering
	\includegraphics[width=0.7\linewidth]{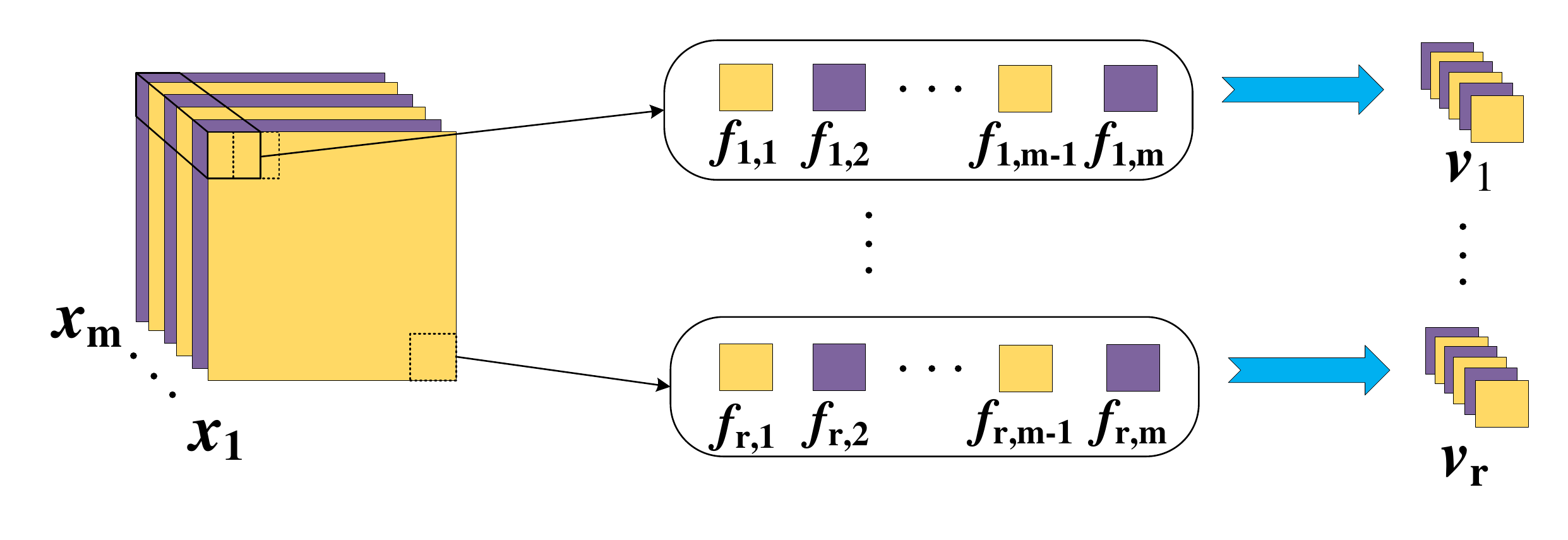}
	\caption{The relationships among  $(\bm{x}_1, \bm{x}_2,\cdots, \bm{x}_r)$, $(\bm{v}_1, \bm{v}_2,\cdots, \bm{v}_r)$, and $(\bm{f}_{1,1},  \bm{f}_{1,2},\cdots, \bm{f}_{r,m}))$. Feature vectors $\bm{f}_{1,1}$ and $\bm{f}_{1,2}$ are in group $\bm{v}_1$, since they are the first feature vectors of different instances.}
	\label{fig:figure3}
\end{figure}

The relationships between these vectors are illustrated in Figure \ref{fig:figure3}. For example, $\bm{f}_{1,1}$ and $\bm{f}_{1,2}$ are members of group $\bm{v}_1$, and are the first feature vectors of instances $\bm{x}_1$ and $\bm{x}_2$, respectively.

Step 2: Obtain binary sequences for all feature vectors using aHash. For example, $(b_{r,i}(k_1), b_{r,i}(k_2),\\\dots, b_{r,i}(k_c))$ is the binary sequence of $b_{r,i}$, where $c$ is the length of the binary sequence and $b_{r,i}(k_j)$ is obtained by aHash. 

Step 3: Calculate the distances between the features in each group. In group $\bm{v}_r$, the $j$-th feature distance $\bm{v}_r(k_j)$ can be calculated using Equation (4):
\begin{equation}
	\bm{v}_r(k_j)=\left\{
	\begin{aligned}
		\frac{\sum_{i=1}^{m} b_{r,i}(k_j)}{m},\quad \frac{\sum_{i=1}^{m} b_{r,i}(k_j)}{m}\textless 0.5\\
		1-\frac{\sum_{i=1}^{m} b_{r,i}(k_j)}{m},\frac{\sum_{i=1}^{m} b_{r,i}(k_j)}{m}\geq 0.5\\
	\end{aligned}
	\right.
\end{equation}

Step 4: Calculate the distance between each pair of groups. This distance represents the differences in the feature vectors in the groups. The larger the distance, the greater the difference in the feature vectors. The distance $d_r$ of group $\bm{v}_r$ can be calculated using Equation (5): 

\begin{equation}
	d_r = \frac{\sum_{j=1}^{c} \bm{v}_r(k_j)}{c}
\end{equation}

Step 5: Rank the groups in descending order based on their distances. We then obtain a new instances order $(\bm{w}_1,\bm{w}_2,\cdots,\bm{w}_r)$ and their distances $(e_1,e_2,\cdots,e_r)$, where $e_1$ is the largest.

Step 6: Calculate the number of groups contained in each percentage of $D$, where $D$ is the sum of all distances, i.e., $D=\sum_{i=1}^{r} e_i$ . We then obtain $(N(100),N(99),\cdots,N(1))$ based on $(e_1,e_2,\cdots,e_r)$. The calculation of $N(g)$ is carried out using Equation (6):

\begin{equation}
	N(g)=\{m|\sum_{i=1}^{m} e_i\geq \frac{Dg}{100}\}
\end{equation}

Step 7: Determine the threshold $HT$. We compare $r/50$ with $N(100)-N(99), N(99)-N(98), \cdots, N(2)-N(1)$ in order. If  $N(p)-N(p-1)$ is the first term that is lower than or equal to $r/50$, we utilize $e_h$ as the threshold $HT$, where $h$ is the value of $N(p)$ and $p\in\{100,99,\cdots,2\}$. $HT$ is calculated using Equation (7):

\begin{equation}
	HT = e_{N(p)}
\end{equation}
where $p=max\{u|N(u)-N(u-1)\geq \frac{r}{100}, u=100,99,\dots,2 \}$.

Step 8: Compare the distances with $HT$. If the distance of a group is lower than the threshold, we eliminate the feature vectors in the group; otherwise, the feature vectors in the group are converted to class vectors and input into the cascade structure.

To illustrate the above steps, we present an example below. 

\begin{figure}
	\centering
	\includegraphics[width=0.7\linewidth]{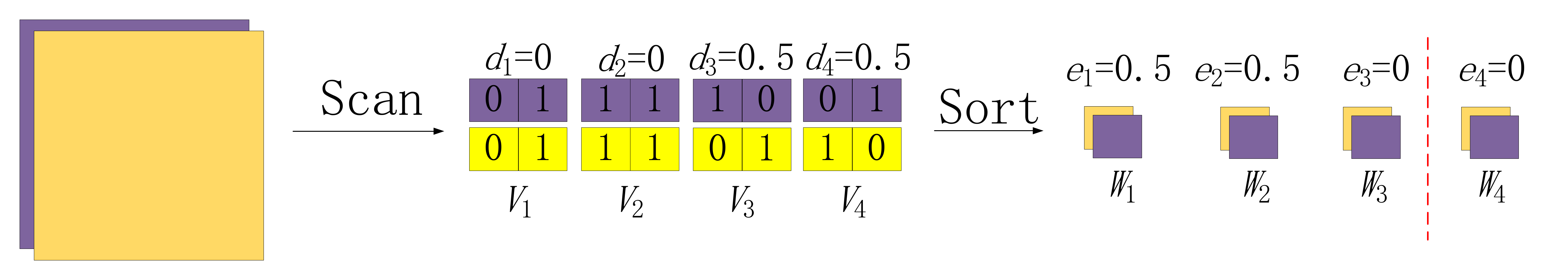}
	\caption{
		Illustration of hashing screening. We select the distance of group 2 as the threshold, since $N(100)-N(99)$ is zero and is therefore lower than $\frac{4}{50}$.}
	\label{fig:figure4}
\end{figure}

\textbf{Example 1}. Suppose we have four groups $(\bm{v}_1, \bm{v}_2,\bm{v}_3, \bm{v}_4)$, for which the feature vectors are $(\bm{f}_{1,1}=(0,1)$,  $\bm{f}_{1,2}=(0,1))$, ($\bm{f}_{2,1}=(1,1)$,  $\bm{f}_{2,2}=(1,1))$, $(\bm{f}_{3,1}=(1,0),  \bm{f}_{3,2}=(0,1))$, $(\bm{f}_{4,1}=(0,1),  \bm{f}_{4,2}=(1,0))$. We know that the values of $\bm{f}_{1,1}(k_1)$ and $\bm{f}_{1,2}(k_1)$ are zero. Hence, according to Equation (4), the value of $\bm{v}_1 (\bm{k}_1 )=\frac{0+0}{2}=0$. Similarly, the values of ($\bm{v}_1 (k_1)$, $\bm{v}_1 (k_2)$), ($\bm{v}_2 (k_1)$, $\bm{v}_2 (k_2)$), ($\bm{v}_3 (k_1)$, $\bm{v}_3 (k_2)$), and ($\bm{v}_4 (k_1)$, $\bm{v}_4 (k_2)$) are (0, 0), (0, 0), (0.5, 0.5), and (0.5, 0.5), respectively. Since $\bm{v}_4 (k_1 )=0.5$ and $\bm{v}_4 (k_2 )=0.5$, the distance of $\bm{v}_4$ is 0.5, i.e., $d_4$=$\frac{0.5+0.5}{2}=0.5$. Thus, $(d_1, d_2, d_3,  d_4)$ are (0, 0, 0.5, 0.5). We then rank the groups according to their distances.
Finally, we obtain ($\bm{w}_1$, $\bm{w}_2$, $\bm{w}_3$, $\bm{w}_4$), for which the values are $(e_1=0.5, e_2=0.5, e_3=0, e_4=0)$. Hence, $D=\sum_{i=1}^{4}e_i=0.5+0.5+0+0=1$. We now obtain the values of $(N(100),N(99),\cdots,N(1))$. According to Equation (6), the values of $N(100), N(99),\dots, N(51)$ are 2, since $e_1+e_2\geq 100\times\frac{D}{100}=1$, and $e_1+e_2\geq 51\times \frac{D}{100}=0.51$. 
Similarly, the values of $N(50), N(49), \cdots, N(1)$ are 1, since $e_1\geq 50\times \frac{D}{100}=0.50$ and $e_1\geq 1\times \frac{D}{100}=0.01$. We know that there are four groups, meaning that $r=4$, and $\frac{r}{50}=\frac{4}{50}=0.08$. According to Step 7, $p=100$, since  $N(100)-N(99)$ is 0, and this is the first value lower than 0.08. Since $p=100$ and $N(100)=2$, according to Equation (7), we select the second group distance $e_2$ as the threshold, i.e., $h=2$. Finally, we eliminate $\bm{w}_3$ and $\bm{w}_4$, since their distances $e_3$ and $e_4$ are lower than $e_2$. $\bm{w}_1$ and $\bm{w}_2$ are converted to class vectors, and are input into the cascade structure. Figure \ref{fig:figure4} illustrates the hashing screening mechanism.

Example 1 verifies that hashing screening can effectively eliminate the redundant feature vectors.

\subsection{Window Screening}

As an improved version of gcForestcs, DBC-Forest avoids the mispartition of instances and thus improves the performance. However, its performance depends on the hyperparameter $k$, which represents the number of bins used to partition instances. Specifically, the smaller the value of $k$, the larger the threshold. Example 2 illustrates this phenomenon.

\begin{figure}
	\centering
	\includegraphics[width=0.7\linewidth]{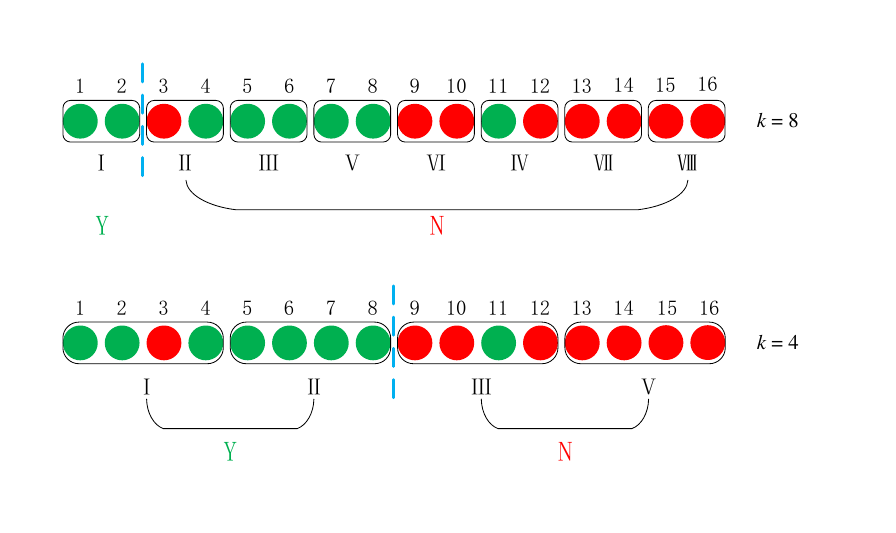}
	\caption{
		Comparison of the different thresholds of DBC-Forest for different values of the hyperparameter $k$. Suppose $TA$ is 70\%. DBC-Forest selects instance 2 as the threshold when $k = 8$, since the accuracy of bin \uppercase\expandafter{\romannumeral2} is 0.5, which is lower than $TA$, and selects instance 8 as threshold when $k = 4$, since the accuracy of bin \uppercase\expandafter{\romannumeral3} is 0.25, which is lower than $TA$.}
	\label{fig:figure5}
\end{figure}

\textbf{Example 2}. Suppose we have 16 instances sorted by confidence, as illustrated in Figure \ref{fig:figure5}. Of these, instances 1, 2, 4, 5, 6, 7, 8, and 11 are correct predictions, and the others are incorrect. We set two cases with $k$ = 8 and $k$ = 4 under the condition of the hyperparameter $TA$ = 70\%. In the case of $k$ = 8, instance 2 is selected as the threshold in DBC-Forest, since the accuracy of bin \uppercase\expandafter{\romannumeral2} is 50\%. In the case of $k$ = 4, instance 8 is selected, since the accuracy of bin \uppercase\expandafter{\romannumeral3} is 25\%. This example shows that the performance of DBC-Forest depends on the hyperparameter $k$. In other words, DBC-Forest cannot achieve high accuracy if the hyperparameter $k$ is not suitable for the dataset. 

To address this drawback of DBC-Forest, we propose a window screening mechanism, in which a self-adaptive approach is used to calculate the threshold. Unlike the fixed window used in DBC-Forest, this approach uses windows with variable size. The mechanism can be summarized as follows.

Step 1: Rank the instances based on their confidences, and obtain $(\bm{x}_1,\bm{x}_2,\cdots,\bm{x}_m )$. 

Step 2: Establish a window $\bm{W}$ for which the parameters are $c$, $u$, and $l$, where $c$ is the size of the window, and $u$ and $l$ are the starting and ending points of the window, respectively. Initially, the values of $c$, $u$, and $l$ are $m/2$, 1, and $m/2$, respectively. The window $\bm{W}$ can be defined according to Equation (8):

\begin{equation}
	(\bm{x}_{u},\bm{x}_{u+1},\dots,\bm{x}_{l})\in\bm{W}
\end{equation}

Step 3: Slide the window $\bm{W}$ using a step size of one (i.e., increase the values of $u$ and $l$), and calculate the accuracy of window $P_{\bm{w}}$ using Equation (9):

\begin{equation}
	P_{\bm{w}} =\frac{\sum_{i=u}^{l}\mathds{1}[p(\bm{x}_{i})=\bm{y}_{i}]}{l-u}
\end{equation}
where $p(\bm{x}_i)$ is the prediction $\bm{x}_i$, and $\bm{y}_i$ is the label of $\bm{x}_i$. If the accuracy is larger than the target accuracy $TA$ and the value of $c$ is not two, the window stops sliding.

Step 4: Reset the window parameters: the value of $c$ is reduced by half, $u$ does not change, and $l=u+c-1$.

Step 5: Iterate Steps 3 and 4. Set the first instance in the window as the threshold $WT$ when $c$ is equal to or lower than two.

Step 6: Compare the confidence values of the instances with the threshold. If the confidence of the instance is larger than the threshold, then the instance produces a final prediction at the current level; otherwise, the instance is passed to the next level.

Unlike in the binning confidence screening approach, the threshold in window screening can be obtained by a self-adaptive method. Moreover, the performance of this approach is better than DBC-Forest, since the bin size hyperparameter is not adjusted. An illustrative example is given below.

\textbf{Example 3}. In this example, we set the target accuracy $TA$ to 70\%, and employ the same 16 instances as in Example 2. The process of window screening is shown in Figure 6. In stage 1, we set the initial values to $c = 8$, $u = 1$, and $l = 8$. We know that the accuracy of the window is 0.875, which is greater than $TA$. Thus, we slide the window $\bm{W}$ using a step size of one. We know that when $u = 2$ and $l = 9$, the accuracy of the window is also 0.875. However, when $u = 3$ and $l = 10$, the accuracy of the window is 0.625, which is less than $TA$. We then enter stage 2, and reset the values to $c = 4$, $u = 3$, and $l = 6$. We know that the accuracy of this window is 0.75, which is greater than $TA$. We then iterate this process until $c = 4$, $u = 7$, and $l = 10$. In this case, the accuracy of the window is 0.50, which is less than $TA$. We then enter stage 3, and reset the values to $c = 2$, $u = 7$, and $l = 8$. In this case, the accuracy of the window is 1.0, which is greater than $TA$. We iterate this process until $c = 2$, $u = 8$, and $l = 9$. In this case, the accuracy of the window is 0.50, which is less than $TA$. Since $c = 2$, the window does not shrunk further. Hence, the threshold is the confidence of instance 8, which is the same as for the case of $k = 4$ in Example 2.

\begin{figure}
	\centering
	\includegraphics[width=0.7\linewidth]{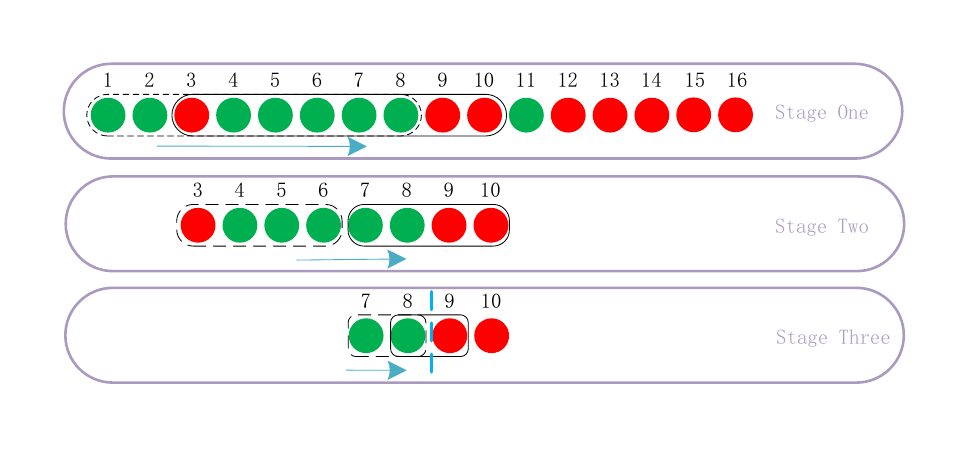}
	\caption{
		Illustration of window screening. The size of the window $c$ will be decreased when the accuracy of the window is lower than $TA$. When the size of window is two and the accuracy is lower than $TA$, the confidence of the first instance is selected as the threshold. }
	\label{fig:figure6}
\end{figure}
\subsection{HW-Forest}

\begin{figure*}
	\centering
	\includegraphics[width=1.0\linewidth]{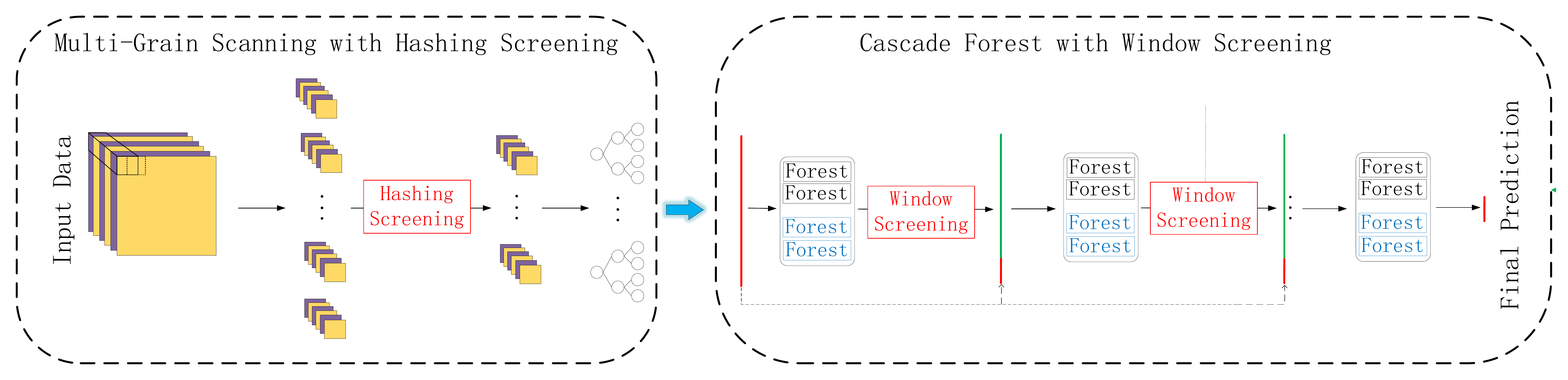}
	\caption{Framework of HW-Forest. A hashing screening strategy is adopted to eliminate redundant feature vectors, and window screening is applied to select high-confidence instances in the cascade forest.}
	\label{fig:figure7}
\end{figure*}

HW-Forest is an improved version of gcForest which consists of two parts: multi-grained scanning with hashing screening and cascade structure with window screening. Figure \ref{fig:figure7} illustrates the overall procedure of HW-Forest. In the first part, the instances are input into the multi-grained scanning stage. We then apply hashing screening to eliminate the redundant feature vectors, and the remaining feature vectors are converted into class vectors using a random forest algorithm. The second part is a cascade forest structure with window screening mechanism to reduce the time costs and memory requirements. The class vectors in the form of the original data are input into the first level. A window screening mechanism is used to divide all instances into high-confidence and low-confidence instances at each level. High-confidence instances are used to produce the final prediction at the current level, while low-confidence instances are passed to the next level.

Algorithm 1 shows pseudocode for HW-Forest.

\begin{algorithm}[htb]
	\label{Algorithm 1}
	\caption{HW-Forest}
	\begin{algorithmic}[1]
		\REQUIRE Training set $\bm{S}$, validation set $\bm{S}_v$, and the maximal number of cascade levels $\bm{T}$.
		\ENSURE HW-Forest model $f$.
		\STATE Get $(\bm{v}_1, \bm{v}_2,\cdots, \bm{v}_r)$ by scanning $\bm{S}$;
		\STATE Calculate $HT$ using a hashing screening mechanism;
		\STATE Eliminate redundant feature vectors and obtain a new training set $S’$;
		\STATE $t=1$, $l_1=0$, $S_1=S’$;
		\WHILE{$t\leq \bm{T}$}
		\STATE $g=99, c=crad(S)/2, u=0, l=c,$ and $l_0=1$;
		\STATE Calculate $WT_t$ using a window screening mechanism;
		\STATE $f_t=prediction(S’)$;
		\STATE Compute the validation error $l_t= f_t (\bm{S_v})$;
		\IF {$l_t>l_(t-1)$}
		\RETURN $f$;
		\ENDIF
		\STATE $f=f_t$
		\STATE $S_{t+1}=S_t \backslash \{ x| x>WT_t, x\in S_t\}$
		\STATE $t=t+1$
		\ENDWHILE
		\RETURN $f$;
	\end{algorithmic}
\end{algorithm}

\section{Experiments}

In this section, we report the results of some experiments carried out to assess the performance of HW-Forest. We also demonstrate the effects of the hashing screening and window screening processes. This section is organized as follows. Section 5.1 introduces the experimental setting, including the parameters, experimental environment, and baseline methods. Section 5.2 discusses the effect of hashing screening, and Section 5.3 illustrates the process of hashing screening by showing instances processed with this method. Section 5.4 discusses the influence of window screening, and Section 5.5 compares the thresholds used in DBC-Forest and HW-Forest to validate the effect of window screening. Section 5.6 compares HW-Forest with other state-of-the-art models on benchmark datasets. 

\subsection{Experimental Setup}

\textbf{Parameter Settings}. In all of our experiments, the models used the same parameters, which were as follows. For the cascade structure, each level was produced by five-fold cross-validation and had a random forest and a completely-random forest. Each forest contained 50 decision trees. The number of cascade levels stopped increasing when the current level did not give an improvement in the accuracy of the previous level. For multi-grained scanning, we established three windows to scan the high-dimensional data, with sizes of 4×4, 6×6, and, 8×8. Each window had a random forest and a completely-random forest, and the number of decision trees in these two forests was 30. $TA$ was set to decrease the error rate by 1/2. 

\textbf{Datasets}. We used nine benchmark datasets to verify the performance of HW-Forest. Table \ref{table1} summarizes the statistics of these datasets. Of these, MNIST, EMNIST, FASHION-MNIST, and QMNIST were commonly used high-dimensional datasets  \cite{wang2021tkd,liu2020pami,zeng2021tkd}, while ADULT, BANK, YEAST, LETTER, IMDB were low-dimensional datasets. 

\begin{table}[]
	\footnotesize
	\caption{Summary of datasets}
	\centering
	\label{table1}
	\begin{tabular}{@{}ccccc@{}}
		\toprule
		Name                    & Training   & Testing   & Features & Labels \\ \midrule
		MNIST                   & 56,000  & 14,000 & 784      & 10     \\
		EMNIST                  & 105,280 & 26,320 & 784      & 10     \\
		FASHION-MNIST           & 56,000  & 14,000 & 784      & 10     \\
		QMNIST                  & 96,000  & 24,000 & 784      & 10     \\
		ADULT                   & 39,074  & 9,768  & 14       & 2      \\
		BANK (BANK   MARKETING) & 32,950  & 8,238  & 20       & 2      \\
		YEAST                   & 1,187   & 297    & 8        & 10     \\
		LETTER                  & 16,000  & 4,000  & 16       & 26     \\
		IMDB                    & 40,000  & 10,000 & 5,000    & 2      \\ \bottomrule
	\end{tabular}
\end{table}

\textbf{Evaluation metrics and methodology}. In all of our experiments, we used the predictive accuracy as a measure of the classification performance, as it is suitable for balanced datasets. The training time was used to evaluate the efficiency. Statistical tests were also adopted as measurement criteria.

\textbf{Hardware}. In these experiments, we used a machine with 2×2.20 GHz CPUs (10 cores) and 128 GB main memory.

\textbf{Baseline methods}. To validate the performance of HW-Forest, we selected four state-of-the-art algorithms as competitive models: gcForest \cite{zhou2019nsr}, gcForestcs \cite{pang2020TKD}, DBC-Forest \cite{ma2021neu}, and AWDF \cite{utk2020ted}. To verify the effect of our hashing screening mechanism, we introduced this mechanism into gcForest, gcForestcs, DBC-Forest, and AWDF, to give H-gcForest, H-gcForestcs, HDBC-Forest, and H-AWDF, respectively. 

\textbf{gcForest} \cite{zhou2019nsr}: This is a deep learning model that is inspired by deep neural networks, and adopts multi-grained scanning and a cascade structure to process the data.

\textbf{gcForestcs} \cite{pang2020TKD}: This is an improved gcForest model in which a confidence screening mechanism is introduced to screen the instances at each level.

\textbf{DBC-Forest} \cite{ma2021neu}: This is an improved gcForestcs model. To increase the accuracy of gcForestcs, binning confidence screening is used to avoid the mispartitioning of instances at each level.

\textbf{AWDF} \cite{utk2020ted}: This is an adaptive weighted model in which adaptive weighting is used for every training instance at each level.

\textbf{H-gcForest, H-gcForestcs, HDBC-Forest, and H-AWDF}: To verify the influence of the hashing screening mechanism, we introduced it into gcForest, gcForestcs, DBC-Forest, and AWDF, to give H-gcForest, H-gcForestcs, HDBC-Forest, and H-AWDF, respectively.

\textbf{H-Forest (H-gcForest) and W-Forest}: To validate the influence of two screening mechanisms, we developed H-Forest (H-gcForest) and W-Forest. These are versions of HW-Forest without the window and hashing screening mechanisms, respectively.

To validate the performance of HW-Forest, we developed the following five research questions (RQ).

RQ 1: What is the effect of hashing screening, and can it be adopted in other models?

RQ 2: What is the effect of hashing screening with different window sizes?

RQ 3: Compared with DBC-Forest, what is the effect of HW-Forest at each level?

RQ 4: What is the difference between the thresholds in DBC-Forest and HW-Forest?

RQ 5: Compared with state-of-the-art alternatives, what is the performance of HW-Forest?

To answer RQ 1, we introduced hashing screening into gcForest, gcForestcs, DBC-Forest, and AWDF, and compared their accuracies with the original models to verify the influence of hashing screening. To address RQ 2, we selected 10 instances from FASHION-MINST to investigate the effect of hashing screening with different window sizes, and to explore the performance of HW-Forest with different window sizes. In response to RQ 3, we compared the accuracy and the number of instances between DBC-Forest and HW-Forest at each level. Specifically, we used MNIST, EMNIST, FASHION-MNIST, QMNIST, and IMDB in this experiment. To answer RQ 4, we explored the thresholds used in the window screening and binning confidence screening mechanisms to show the difference. For RQ 5, we compared the performance of HW-Forest with gcForest, gcForestcs, DBC-Forest, AWDF, H-Forest, and W-Forest.

\subsection{Influence of Hashing Screening}

To analyze the effect of the hashing screening mechanism, we conducted four experiments to compare the performance between pairs of models: gcForest and H-gcForest; gcForestcs and H-gcForestcs; gcForestcs and H-gcForestcs; and AWDF and H-AWDF. Experiments were conducted on the MNIST, EMNIST, FASHION-MNIST, and QMNIST datasets, since these need to be processed by multi-grained scanning. The results of this comparison are shown in Tables \ref{table2} to \ref{table6}.

\begin{table*}[]
	\caption{Comparison of performance between gcForest and H-gcForest}
		\centering
	\footnotesize
	\label{table2}
	\begin{tabular}{@{}
			>{\columncolor[HTML]{FFFFFF}}c 
			>{\columncolor[HTML]{FFFFFF}}c 
			>{\columncolor[HTML]{FFFFFF}}c 
			>{\columncolor[HTML]{FFFFFF}}c 
			>{\columncolor[HTML]{FFFFFF}}c 
			>{\columncolor[HTML]{FFFFFF}}c 
			>{\columncolor[HTML]{FFFFFF}}c @{}}
		\toprule
		\multicolumn{1}{c}{\cellcolor[HTML]{FFFFFF}}                            & \multicolumn{3}{c}{\cellcolor[HTML]{FFFFFF}Accuracy   (\%)} & \multicolumn{3}{c}{\cellcolor[HTML]{FFFFFF}Time(s)}  \\  \cmidrule(l){2-7}
		\multicolumn{1}{c}{\multirow{-2}{*}{\cellcolor[HTML]{FFFFFF}Dataset}} & gcForest           & H-gcForest          & \multicolumn{1}{c}{\cellcolor[HTML]{FFFFFF}Difference}        & gcForest       & H-gcForest    & Difference          \\   \midrule
		MNIST                                                                    & \textbf{98.77±0.17}         & 98.75±0.12         & -0.02             & 1972.27±6.39   & \textbf{1464.86±11.54} & 507.41 (34.70\%)    \\
		EMNIST                                                                   & \textbf{86.18±0.24}         & 86.13±0.28         & -0.05             & 8774.77±108.08 & \textbf{6920.12±51.39} & 1854.65   (26.80\%) \\
		FASHION-MNIST                                                            & 89.94±0.29         & \textbf{90.05±0.29}         & 0.09              & 2411.10±10.28  & \textbf{2272.63±25.18} & 138.47 (6.09\%)     \\
		QMNIST                                                                   & \textbf{98.93±0.06}         & 98.92±0.06         & -0.01             & 2332.79±31.80  & \textbf{1790.10±17.71} & 542.69   (30.32\%)  \\  \bottomrule
	\end{tabular}
\end{table*}

\begin{table*}[]
	\caption{Comparison of performance between gcForestcs and H-gcForestcs}
		\centering
	\footnotesize
	\label{table3}
	\begin{tabular}{@{}
			>{\columncolor[HTML]{FFFFFF}}c 
			>{\columncolor[HTML]{FFFFFF}}c
			>{\columncolor[HTML]{FFFFFF}}c 
			>{\columncolor[HTML]{FFFFFF}}c 
			>{\columncolor[HTML]{FFFFFF}}c 
			>{\columncolor[HTML]{FFFFFF}}c 
			>{\columncolor[HTML]{FFFFFF}}c @{}}
		\toprule
		\multicolumn{1}{c}{\cellcolor[HTML]{FFFFFF}}                           & \multicolumn{3}{c}{\cellcolor[HTML]{FFFFFF}Accuracy   (\%)} & \multicolumn{3}{c}{\cellcolor[HTML]{FFFFFF}Time(s)} \\ \cmidrule(l){2-7}
		\multicolumn{1}{c}{\multirow{-2}{*}{\cellcolor[HTML]{FFFFFF}Dataset}} & gcForestcs        & H-gcForestcs         & \multicolumn{1}{c}{\cellcolor[HTML]{FFFFFF}Difference}        & gcForestcs     & H-gcForestcs   & Difference        \\ \midrule
		MNIST                                                                   & \textbf{98.20±0.09}        & 98.19±0.12          & -0.01             & 1636.00±13.47  & \textbf{1153.50±8.65}   & 482.5   (41.83\%) \\
		EMNIST                                                                  & 87.11±0.23        & \textbf{87.13±0.30}          & 0.02              & 5945.37±159.68 & \textbf{5313.32±128.73} & 532.05 (11.90\%)  \\
		FASHION-MNIST                                                           & 89.94±0.29        & \textbf{90.03±0.30}          & 0.07              & 1943.09±6.50   & \textbf{1850.41±8.41}   & 92.47   (5.01\%)  \\
		QMNIST                                                                  & 98.40±0.12        & \textbf{98.41±0.12}          & 0.01              & 1991.11±17.71  & \textbf{1454.08±14.77}  & 537.03 (36.93\%)  \\ \bottomrule
	\end{tabular}
\end{table*}

\begin{table*}[]
	\caption{Comparison of performance between DBC-Forest and HDBC-gcForest}
		\centering
	\footnotesize
	\label{table4}
	\begin{tabular}{@{}
			>{\columncolor[HTML]{FFFFFF}}c 
			>{\columncolor[HTML]{FFFFFF}}c 
			>{\columncolor[HTML]{FFFFFF}}c 
			>{\columncolor[HTML]{FFFFFF}}c 
			>{\columncolor[HTML]{FFFFFF}}c 
			>{\columncolor[HTML]{FFFFFF}}c 
			>{\columncolor[HTML]{FFFFFF}}c @{}}
		\toprule
		\multicolumn{1}{c}{\cellcolor[HTML]{FFFFFF}}                           & \multicolumn{3}{c}{\cellcolor[HTML]{FFFFFF}Accuracy   (\%)} & \multicolumn{3}{c}{\cellcolor[HTML]{FFFFFF}Time(s)}  \\ \cmidrule(l){2-7} 
		\multicolumn{1}{c}{\multirow{-2}{*}{\cellcolor[HTML]{FFFFFF}Dataset}} & DBC-Forest         & HDBC-Forest        & \multicolumn{1}{c}{\cellcolor[HTML]{FFFFFF}Difference}        & DBC-Forest     & HDBC-Forest    & Difference         \\  \midrule
		MNIST                                                                   & \textbf{99.03±0.07}         & 99.01±0.06         & -0.02             & 1636.10±17.79  & \textbf{1159.99±9.90}   & 479.11   (41.04\%) \\
		EMNIST                                                                  & 86.82±0.26         & \textbf{86.83±0.16}         & 0.01              & 7226.23±112.34 & \textbf{6349.11±162.52} & 877.12 (13.81\%)   \\
		FASHION-MNIST                                                           & 90.57±0.29         & \textbf{90.70±0.20}         & 0.13              & 2056.95±13.38  & \textbf{1928.22±13.91}  & 128.73   (6.68\%)  \\
		QMNIST                                                                  & \textbf{99.08±0.12}         & 99.05±0.06         & -0.03             & 1979.00±22.14  & \textbf{1467.99±21.23}  & 511.01 (34.81\%)   \\ \bottomrule
	\end{tabular}
\end{table*}

\begin{table*}[]
	\caption{Comparison of performance between AWDF and H-AWDF}
		\centering
	\footnotesize 
	\label{table5}
	\begin{tabular}{@{}
			>{\columncolor[HTML]{FFFFFF}}c 
			>{\columncolor[HTML]{FFFFFF}}c 
			>{\columncolor[HTML]{FFFFFF}}c 
			>{\columncolor[HTML]{FFFFFF}}c 
			>{\columncolor[HTML]{FFFFFF}}c 
			>{\columncolor[HTML]{FFFFFF}}c 
			>{\columncolor[HTML]{FFFFFF}}c @{}}
		\toprule
		\multicolumn{1}{c}{\cellcolor[HTML]{FFFFFF}}                           & \multicolumn{3}{c}{\cellcolor[HTML]{FFFFFF}Accuracy   (\%)} & \multicolumn{3}{c}{\cellcolor[HTML]{FFFFFF}Time(s)}  \\ \cmidrule(l){2-7}
		\multicolumn{1}{c}{\multirow{-2}{*}{\cellcolor[HTML]{FFFFFF}Dataset}} & AWDF               & H-AWDF              & \multicolumn{1}{c}{\cellcolor[HTML]{FFFFFF}Difference}         & AWDF           & H-AWDF         & Difference         \\  \midrule
		MNIST                                                                   & \textbf{98.80±0.11}         & \textbf{98.80±0.11}         & 0.00                 & 1697.79±18.36  & \textbf{1192.30±11.95}  & 505.49   (42.40\%) \\
		EMNIST                                                                  & \textbf{86.74±0.18}         & 86.70±0.25         & -0.04             & 7385.44±103.03 & \textbf{6664.62±201.76} & 720.82 (10.82\%)   \\
		FASHION-MNIST                                                           & 89.89±0.27         & \textbf{90.05±0.14}         & 0.06              & 1821.65±15.30  & \textbf{1734.06±16.83}  & 87.59   (5.05\%)   \\
		QMNIST                                                                  & 98.93±0.08         & \textbf{98.95±0.10}         & 0.02              & 1988.55±66.69  & \textbf{1458.53±14.06}  & 530.02 (36.34\%)   \\ \bottomrule
	\end{tabular}
\end{table*}

\begin{table*}[]
	\caption{Comparison of performance between W-Forest and HW-Forest}
		\centering
	\footnotesize
	\label{table6}
	\begin{tabular}{@{}
			>{\columncolor[HTML]{FFFFFF}}c 
			>{\columncolor[HTML]{FFFFFF}}c 
			>{\columncolor[HTML]{FFFFFF}}c 
			>{\columncolor[HTML]{FFFFFF}}c 
			>{\columncolor[HTML]{FFFFFF}}c 
			>{\columncolor[HTML]{FFFFFF}}c 
			>{\columncolor[HTML]{FFFFFF}}c @{}}
		\toprule
		\multicolumn{1}{c}{\cellcolor[HTML]{FFFFFF}}                           & \multicolumn{3}{c}{\cellcolor[HTML]{FFFFFF}Accuracy   (\%)}                      & \multicolumn{3}{c}{\cellcolor[HTML]{FFFFFF}Time(s)} \\ \cmidrule(l){2-7} 
		\multicolumn{1}{c}{\multirow{-2}{*}{\cellcolor[HTML]{FFFFFF}Dataset}} & W-Forest   & HW-Forest  & \multicolumn{1}{c}{\cellcolor[HTML]{FFFFFF}Difference} & W-Forest       & HW-Forest     & Difference         \\ \midrule
		MNIST                                                                   & 99.05±0.08 & \textbf{99.07±0.09} & -0.02                                                   & 1648.92±20.56  & \textbf{1135.42±9.32}  & 513.5 (45.23\%)    \\
		EMNIST                                                                  & \textbf{87.24±0.18} & 87.22±0.17 & -0.02                                                   & 6856.64±80.16  & \textbf{6408.14±65.33} & 448.5   (7.00\%)   \\
		FASHION-MNIST                                                           & 90.77±0.34 & \textbf{90.79±0.24} & 0.02                                                    & 1990.47±13.91  & \textbf{1904.17±22.82} & 86.30 (4.53\%)     \\
		QMNIST                                                                  & \textbf{99.13±0.05} & \textbf{99.13±0.05} & 0.00                                                       & 1965.98±34.68  & \textbf{1473.09±12.00} & 489.89   (33.46\%) \\ \bottomrule
	\end{tabular}
\end{table*}

The experimental results give rise to the following observations.

1.	Hashing screening effectively decreases the time cost for all datasets. For example, from Table \ref{table3}, it can be seen that H-gcForestcs has a time cost that is 482.5 s (41.83\%) lower than for gcForestcs on the MNIST dataset. This phenomenon can also be seen in the other tables. The reason for this is as follows. We know that the greater the number of feature vectors, the higher the time cost. Hashing screening can effectively eliminate the redundant feature vectors, and hence the time cost is reduced.

2.	Hashing screening does not influence the accuracy of the models. The results of 20 experiments are presented in Tables \ref{table2} to \ref{table6}, and it can be seen that the models with hashing screening have better accuracy than the original models in 11 experiments, worse accuracy than the original models in seven experiments, and the same as the original models in two experiments. For example, Table \ref{table3} shows that H-gcForestcs has better accuracy than gcForestcs on three datasets (EMNIST, FASHION-MNIST, and QMNIST) and worse accuracy than gcForestcs on MNIST. The reason for this is as follows. Since the important feature vectors have larger distances than the redundant feature vectors, hashing screening effectively screens for important feature vectors based on their distances. More importantly, these redundant feature vectors make lower contributions to the classification accuracy. Hence, the models using hashing screening have almost the same accuracy as the original models.

To further evaluate the influence of hashing screening, we applied a $t$-test for statistical hypothesis test.

To measure the influence of the hashing screening mechanism, we applied a $t$-test to the time cost and accuracy. More specifically, we compared the accuracy and time cost of the models at each fold, and obtained the different values $(D_1,D_2,\cdots,D_k)$. We then calculated the statistics according to $\left| \sqrt{k}\mu/\sigma \right|$, where $k$ is the number of cross-validation folds, $\mu$ is the mean of $(D_1,D_2,\cdots,D_k )$, and $\sigma$ is the standard deviation of $(D_1,D_2,\cdots,D_k )$. The results for the time cost and accuracy are shown in Tables \ref{table7} and \ref{table8}, respectively.

\begin{table*}[]
	\footnotesize 
	\caption{Significance testing for time cost}
	\centering
	\label{table7}
	\begin{tabular}{@{}ccccccl@{}}
		\toprule
		\cellcolor[HTML]{FFFFFF}DATASET      & \cellcolor[HTML]{FFFFFF}H-gcForest   & \cellcolor[HTML]{FFFFFF}H-gcForestcs & \cellcolor[HTML]{FFFFFF}HDBC-Forest & \cellcolor[HTML]{FFFFFF}H-AWDF     & \cellcolor[HTML]{FFFFFF}HW-Forest   &  \\ \midrule
		\cellcolor[HTML]{FFFFFF}MNIST         & \cellcolor[HTML]{FFFFFF}R   (192.15) & \cellcolor[HTML]{FFFFFF}R   (52.08)  & \cellcolor[HTML]{FFFFFF}R   (42.58) & \cellcolor[HTML]{FFFFFF}R   (44.3) & \cellcolor[HTML]{FFFFFF}R   (44.30) &  \\
		\cellcolor[HTML]{FFFFFF}EMNIST        & \cellcolor[HTML]{FFFFFF}R (28.02)    & \cellcolor[HTML]{FFFFFF}R (6.56)     & \cellcolor[HTML]{FFFFFF}R (6.83)    & \cellcolor[HTML]{FFFFFF}R (5.99)   & \cellcolor[HTML]{FFFFFF}R (5.36)    &  \\
		\cellcolor[HTML]{FFFFFF}FASHION-MNIST & \cellcolor[HTML]{FFFFFF}R (12.58)    & \cellcolor[HTML]{FFFFFF}R   (35.44)  & \cellcolor[HTML]{FFFFFF}R   (11.49) & \cellcolor[HTML]{FFFFFF}R   (7.71) & \cellcolor[HTML]{FFFFFF}R   (7.71)  &  \\
		\cellcolor[HTML]{FFFFFF}QMNIST        & \cellcolor[HTML]{FFFFFF}R (25.45)    & \cellcolor[HTML]{FFFFFF}R (44.35)    & \cellcolor[HTML]{FFFFFF}R (34.31)   & \cellcolor[HTML]{FFFFFF}R (28.76)  & \cellcolor[HTML]{FFFFFF}R (28.76)   &  \\\bottomrule
	\end{tabular}
\end{table*}

\begin{table*}[]
	\footnotesize 
	\caption{Significance testing for accuracy}
	\centering
	\label{table8}
	\begin{tabular}{@{}ccccccl@{}}
		\toprule
		\cellcolor[HTML]{FFFFFF}DATASET     & \cellcolor[HTML]{FFFFFF}H-gcForest & \cellcolor[HTML]{FFFFFF}H-gcForestcs & \cellcolor[HTML]{FFFFFF}HDBC-Forest & \cellcolor[HTML]{FFFFFF}H-AWDF     & \cellcolor[HTML]{FFFFFF}HW-Forest  &  \\ \midrule
		\cellcolor[HTML]{FFFFFF}MNIST         & \cellcolor[HTML]{FFFFFF}A (0.88)   & \cellcolor[HTML]{FFFFFF}A (1.52)     & \cellcolor[HTML]{FFFFFF}A (0.47)    & \cellcolor[HTML]{FFFFFF}A (1.18)   & \cellcolor[HTML]{FFFFFF}A (0.60)   &  \\
		\cellcolor[HTML]{FFFFFF}EMNIST        & \cellcolor[HTML]{FFFFFF}A   (0.08) & \cellcolor[HTML]{FFFFFF}A   (1.99)   & \cellcolor[HTML]{FFFFFF}A   (0.13)  & \cellcolor[HTML]{FFFFFF}A   (0.98) & \cellcolor[HTML]{FFFFFF}A   (0.75) &  \\
		\cellcolor[HTML]{FFFFFF}FASHION-MNIST & \cellcolor[HTML]{FFFFFF}A (1.15)   & \cellcolor[HTML]{FFFFFF}A (1.89)     & \cellcolor[HTML]{FFFFFF}A (0.80)    & \cellcolor[HTML]{FFFFFF}A (1.17)   & \cellcolor[HTML]{FFFFFF}A (0.09)   &  \\
		\cellcolor[HTML]{FFFFFF}QMNIST        & \cellcolor[HTML]{FFFFFF}A   (0.87) & \cellcolor[HTML]{FFFFFF}A   (0.33)   & \cellcolor[HTML]{FFFFFF}A   (1.13)  & \cellcolor[HTML]{FFFFFF}A   (0.40) & \cellcolor[HTML]{FFFFFF}A   (0.84) &  \\ \bottomrule
	\end{tabular}
\end{table*}

We can reject (R) or accept (A) the null hypothesis based on the statistics. For example, for the MNIST dataset in Table \ref{table8}, the accuracy of H-gcForest is not significantly different from that of gcForest, since $T_{0.05,4}$ is 2.13, which is larger than the $t$-test statistic of 0.88. Furthermore, for the MNIST dataset, Table \ref{table7} shows that the time costs of H-Forest and HW-Forest are significantly different, since the $t$-test statistic is 192.15, which is much larger than 2.13. The results of these statistical hypothesis tests showed that hashing screening effectively reduces the time cost by eliminating redundant feature vectors, without influencing the accuracy. More importantly, when used with other deep forest models, hashing screening can achieve the same accuracy as the original models.

Based on the above experimental results and statistical analysis, we can conclude that hashing screening can reduce the time cost without influencing the accuracy.

\subsection{Effect of Hashing Screening}

\begin{figure}[H]
	\centering
	\subfigure[Ten raw instances.]{ 
		\centering 
		\includegraphics[width=0.8\linewidth]{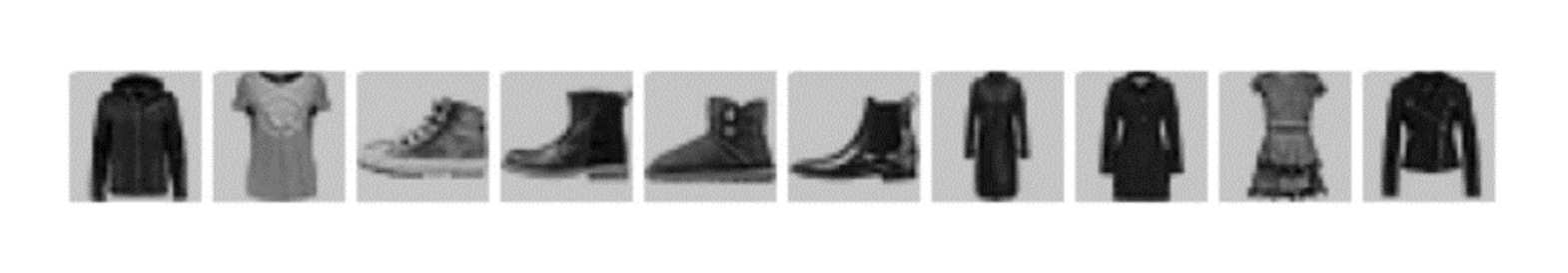}
	}
	\quad
	\subfigure[Processing by hashing screening with a window of size 2×2]{ 
		\centering 
		\includegraphics[width=0.8\linewidth]{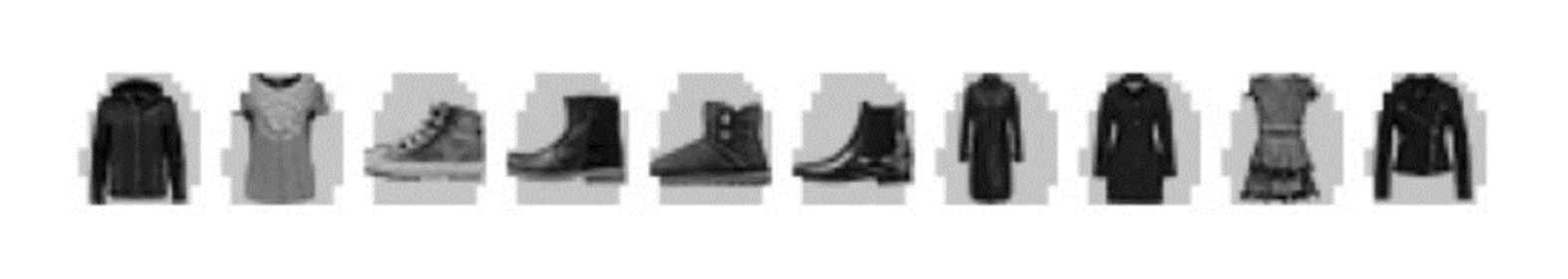}
	}
	\subfigure[Processing by hashing screening with a window of size 4×4]{ 
		\centering 
		\includegraphics[width=0.8\linewidth]{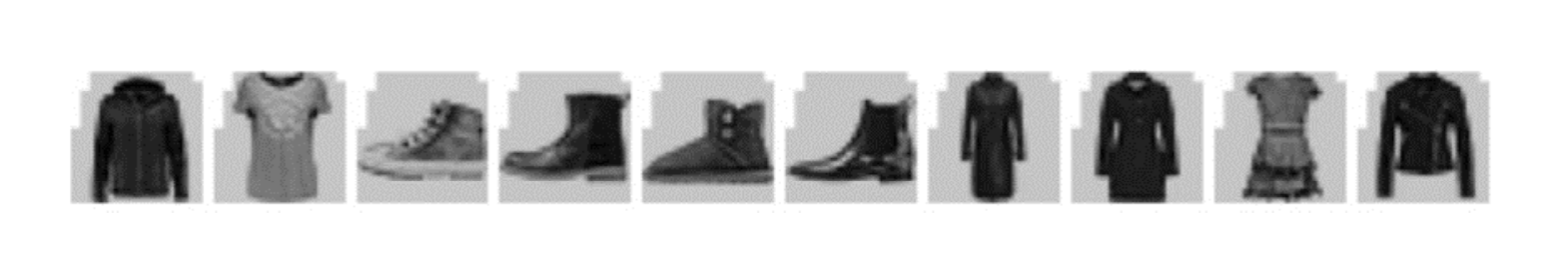}
	}
	\subfigure[Processing by hashing screening with a window of size 6×6]{ 
		\centering 
		\includegraphics[width=0.8\linewidth]{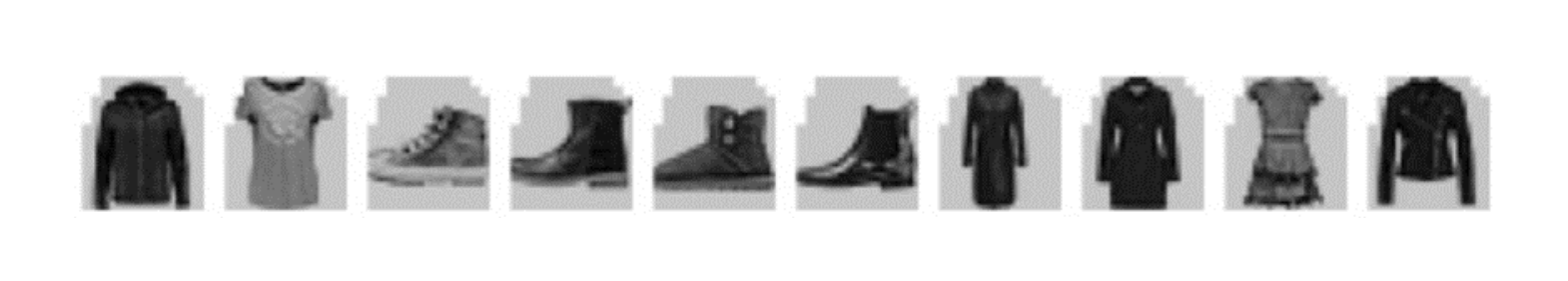}
	}
	\subfigure[Processing by hashing screening with a window of size 8×8]{ 
		\centering 
		\includegraphics[width=0.8\linewidth]{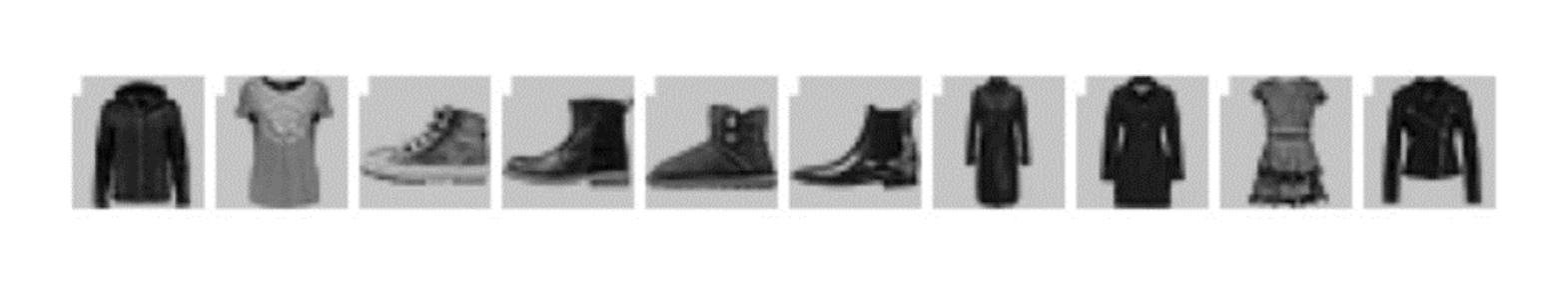}
	}
	\caption{Comparison of raw and processed instances}
	\label{fig:figure8}
\end{figure}

In this section, we compared raw instances with instances processed by hashing screening, to explore the effect of the hashing screening mechanism. We conducted experiments with windows of size 2×2, 4×4, 6×6, and 8×8, and selected 10 instances from the FASHION-MNIST dataset. The results are shown in Figure \ref{fig:figure8}.
Two observations can be made from these results.

1.	The number of redundant feature vectors produced by a large window is significantly lower than for a small window. For example, the number of redundant feature vectors in Figure \ref{fig:figure8}(f) is two, for a window size of 8×8; however, the number of redundant feature vectors in Figure \ref{fig:figure8}(b) is 55, for a window size of 2×2. The reason for this is that on the FASHION-MNSIT dataset, the difference between the number of redundant feature vectors and important feature vectors is small. As the window size increases, the probability of containing important feature vectors increases, and thus the difference becomes smaller. Hence, with an increase in the window size, the number of redundant feature vectors decreases.

2.	Hashing screening reduces the number of redundant feature vectors while retaining important information. For example, as shown in Figure \ref{fig:figure8}(b), hashing screening retains the important features of instances, such as the toes of shoes and the sleeves of coats. The same phenomenon can be seen in the other figures. The reason for this is as follows. Since the important feature vectors are significant different from each other, the distance between them is larger than for the redundant feature vectors. The redundant feature vectors are therefore eliminated and the important feature vectors are retained based on their distances.

To further explore the effect of the hashing screening mechanism, we conducted four experiments with windows of size 2×2, 4×4, 6×6, and 8×8. In each experiment, we compared the accuracy and time cost of W-gcForest and HW-gcForest for different window sizes. The results are shown in Table \ref{table9}.

\begin{table*}[]
	\caption{Comparison of performance between W-Forest and HW-gcForest for different window sizes}
		\centering
	\footnotesize  
	\label{table9}
	\begin{tabular}{
			>{\columncolor[HTML]{FFFFFF}}c 
			>{\columncolor[HTML]{FFFFFF}}c
			>{\columncolor[HTML]{FFFFFF}}c 
			>{\columncolor[HTML]{FFFFFF}}c 
			>{\columncolor[HTML]{FFFFFF}}c 
			>{\columncolor[HTML]{FFFFFF}}c 
			>{\columncolor[HTML]{FFFFFF}}c }
		\toprule
		\multicolumn{1}{c}{\cellcolor[HTML]{FFFFFF}}                           & \multicolumn{3}{c}{\cellcolor[HTML]{FFFFFF}Accuracy   (\%)} & \multicolumn{3}{c}{\cellcolor[HTML]{FFFFFF}Time(s)} \\ \cmidrule(l){2-7}
		\multicolumn{1}{c}{\multirow{-2}{*}{\cellcolor[HTML]{FFFFFF}Window size}} & W-Forest        & HW-Forest         & \multicolumn{1}{c}{\cellcolor[HTML]{FFFFFF}Difference}        & W-Forest     & HW-Forest   & Difference        \\ \hline
		2×2                                                                   & 89.98±0.27        & \textbf{90.05±0.18}          & 0.07             & 989.19±9.25  & \textbf{762.11±10.32}   & 227.08 (29.80\%) \\
		4×4                                                                & 90.57±0.27        & \textbf{90.66±0.10}         & 0.09              & 1,049.75±8.36 & \textbf{953.53±9.68} & 96.22 (10.09\%)  \\
		6×6                                                         & \textbf{90.28±0.28}         & \textbf{90.28±0.28}           & 0.00              & 1,077.05±11.31   & \textbf{1,020.22±11.81}   & 56.83 (5.57\%)  \\
		8×8                                                                 & \textbf{98.98±0.33}         & \textbf{98.98±0.33}           & 0.00              & 1,192.44±11.48  & \textbf{1,184.81±13.19}  & 7.63	(0.64\%)  \\ \hline
	\end{tabular}
\end{table*}

Based on these results, two observations can be made.

1.	With an increase in the window size, the time required for hash screening decreases. As shown in Table \ref{table9}, this time is reduced by 227.08 s (29.80\%) and 56.83 s (5.57\%) when the window sizes are 2×2 and 6×6, respectively. The reason for this is as follows. As in the previous analysis, with an increase in the window size, the number of redundant feature vectors decreases, and hence the time required for hash screening decreases.

2.	Hashing screening does not affect the accuracy of the model with different window sizes. For example, HW-Forest has the same accuracy as W-Forest for window sizes of 8×8 and 6×6, and has better accuracy than W-Forest when the window sizes are 4×4 and 2×2. This phenomenon indicates that using the distance of the feature vector as the criterion for eliminating redundant feature vectors can effectively identify redundant feature vectors with different window sizes. In other words, hashing screening is robust under different window sizes.

\subsection{Influence of Window Screening}

\begin{figure}[H]
	\centering
	\includegraphics[width=0.39\linewidth]{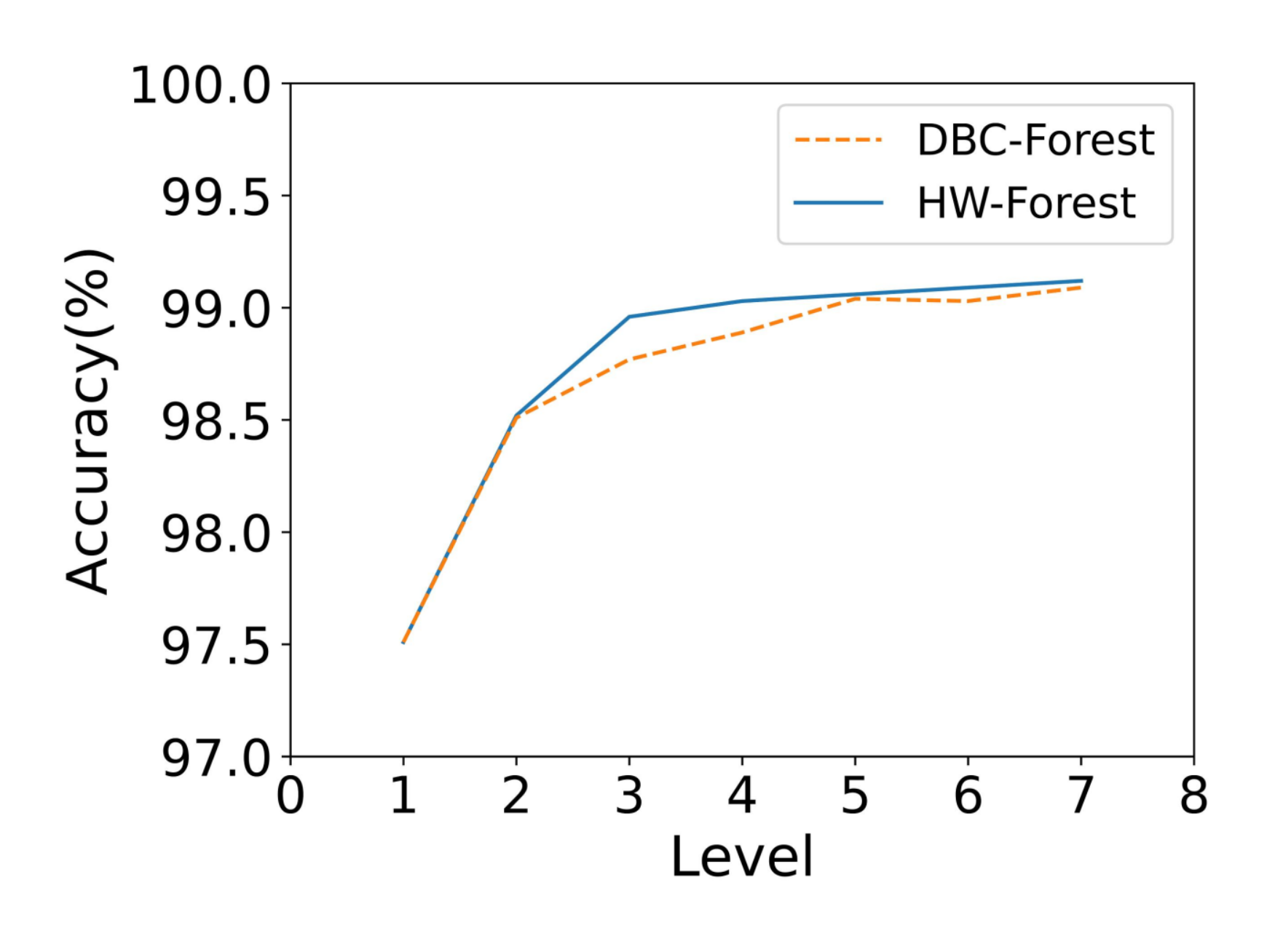}
	\includegraphics[width=0.42\linewidth]{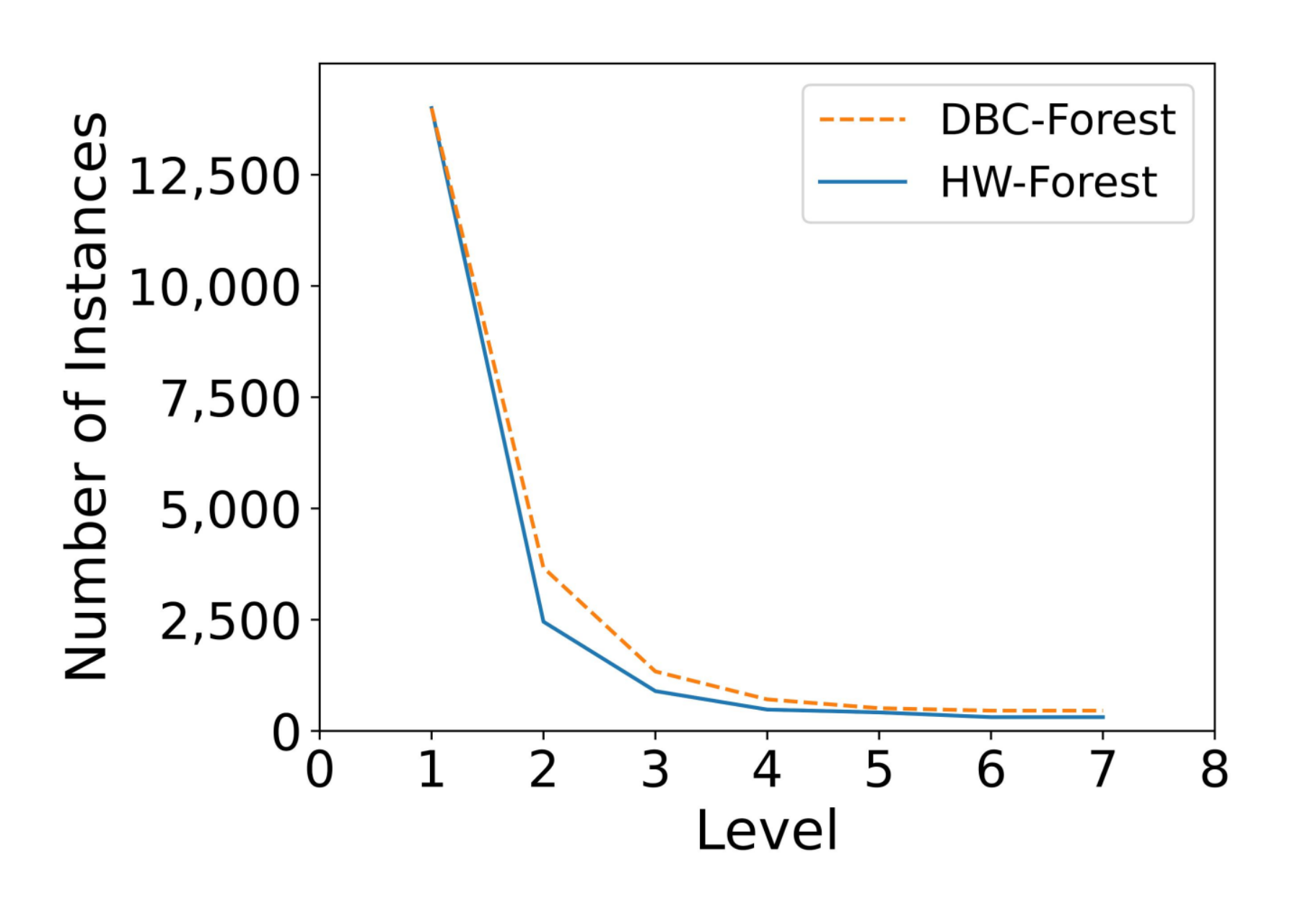}
	
	\caption{Comparison of accuracy and number of instances for the MNIST dataset}
	\label{fig:figure9}
\end{figure}

\begin{figure}[H]
	\centering
	\includegraphics[width=0.39\linewidth]{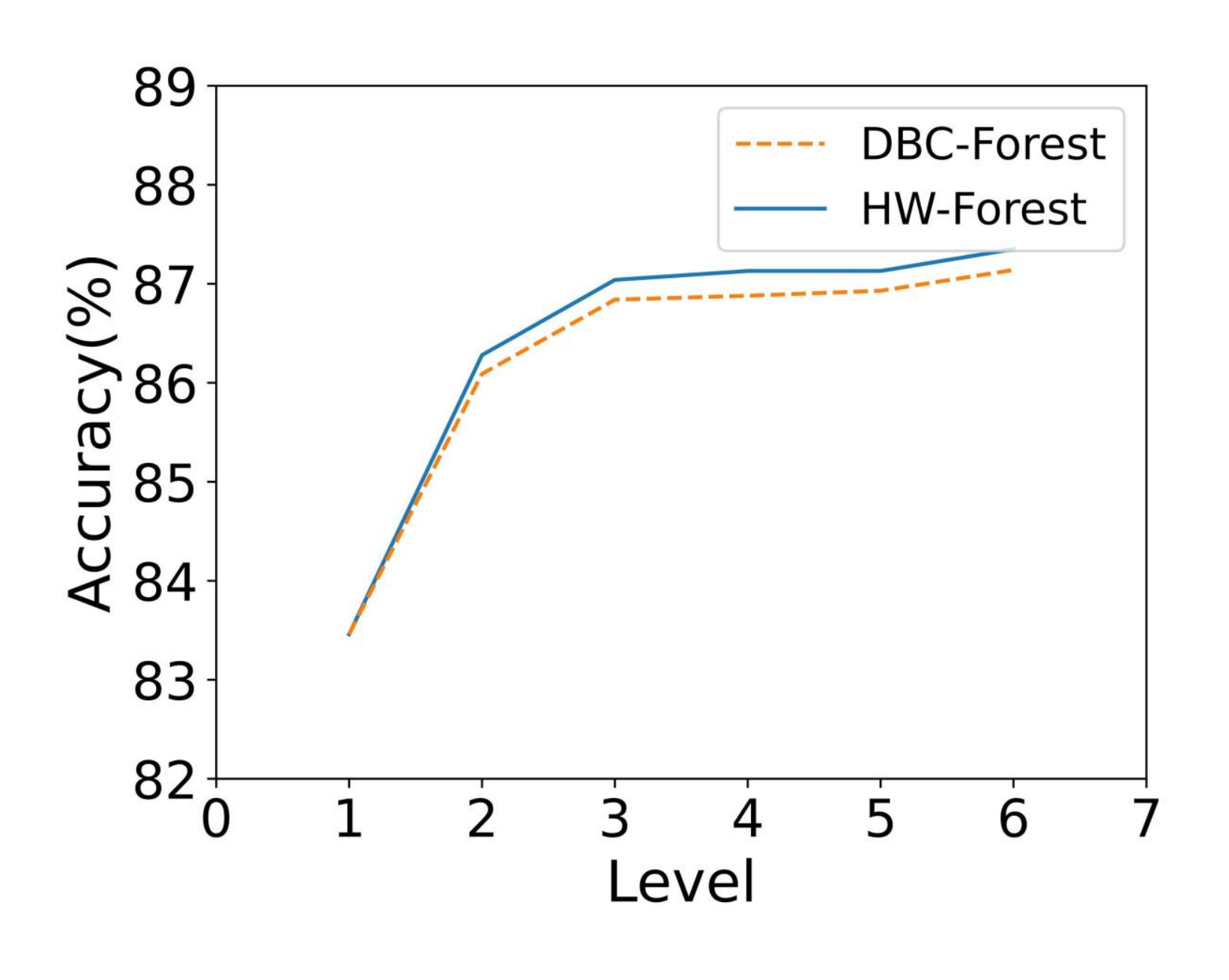}
	\includegraphics[width=0.42\linewidth]{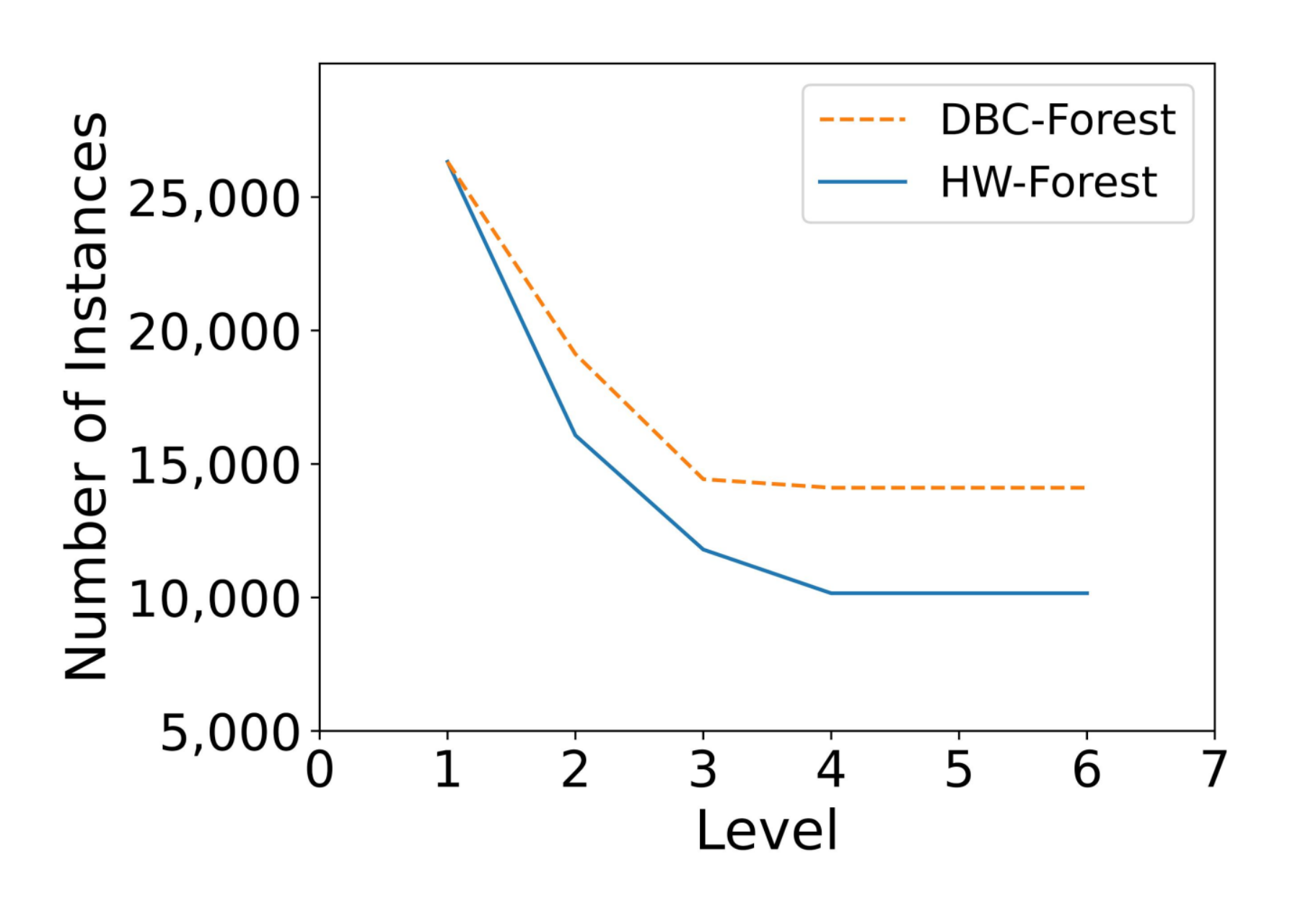}
	
	\caption{Comparison of accuracy and number of instances for the EMNIST dataset}
	\label{fig:figure10}
\end{figure}

\begin{figure}[H]
	\centering
	\includegraphics[width=0.39\linewidth]{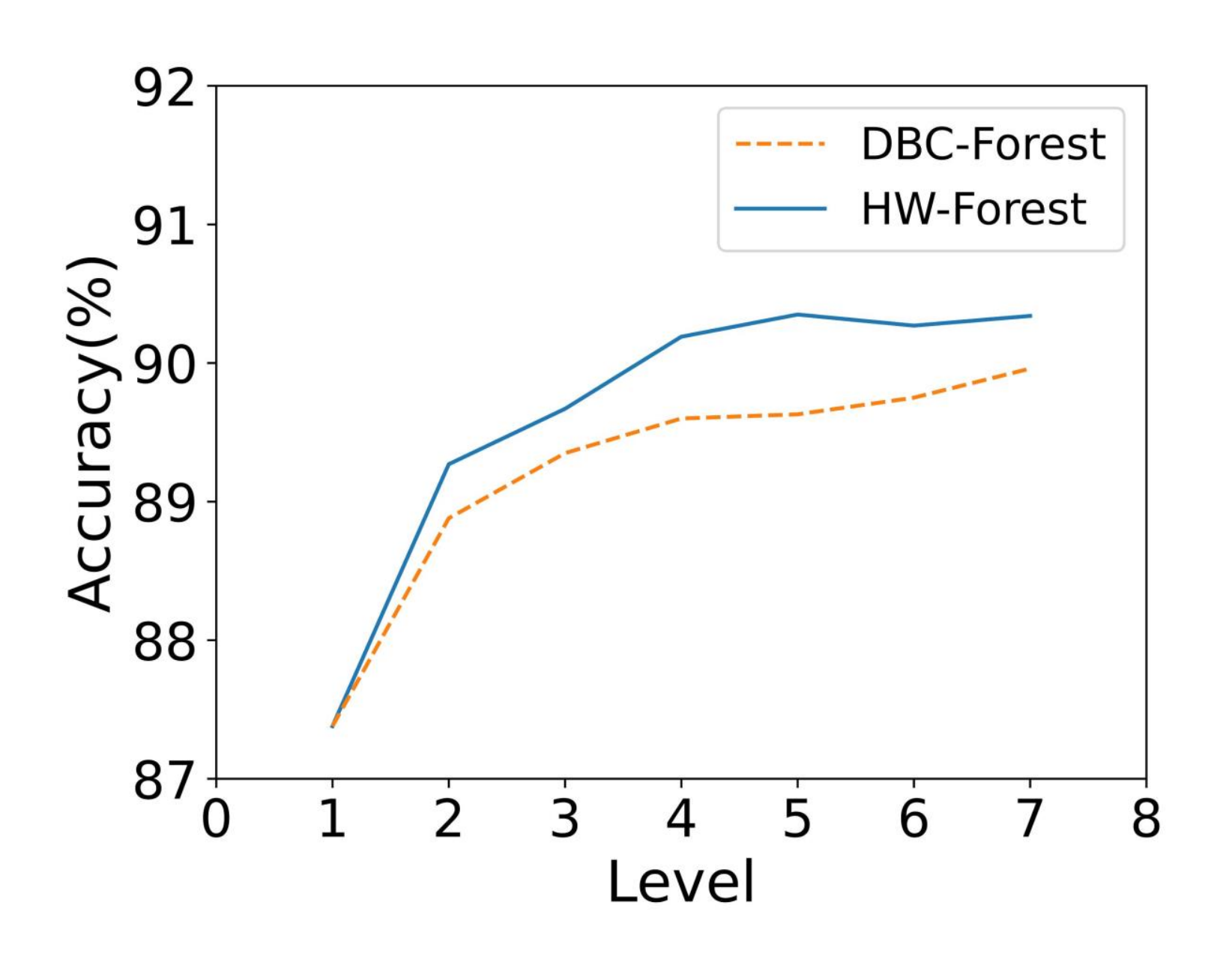}
	\includegraphics[width=0.42\linewidth]{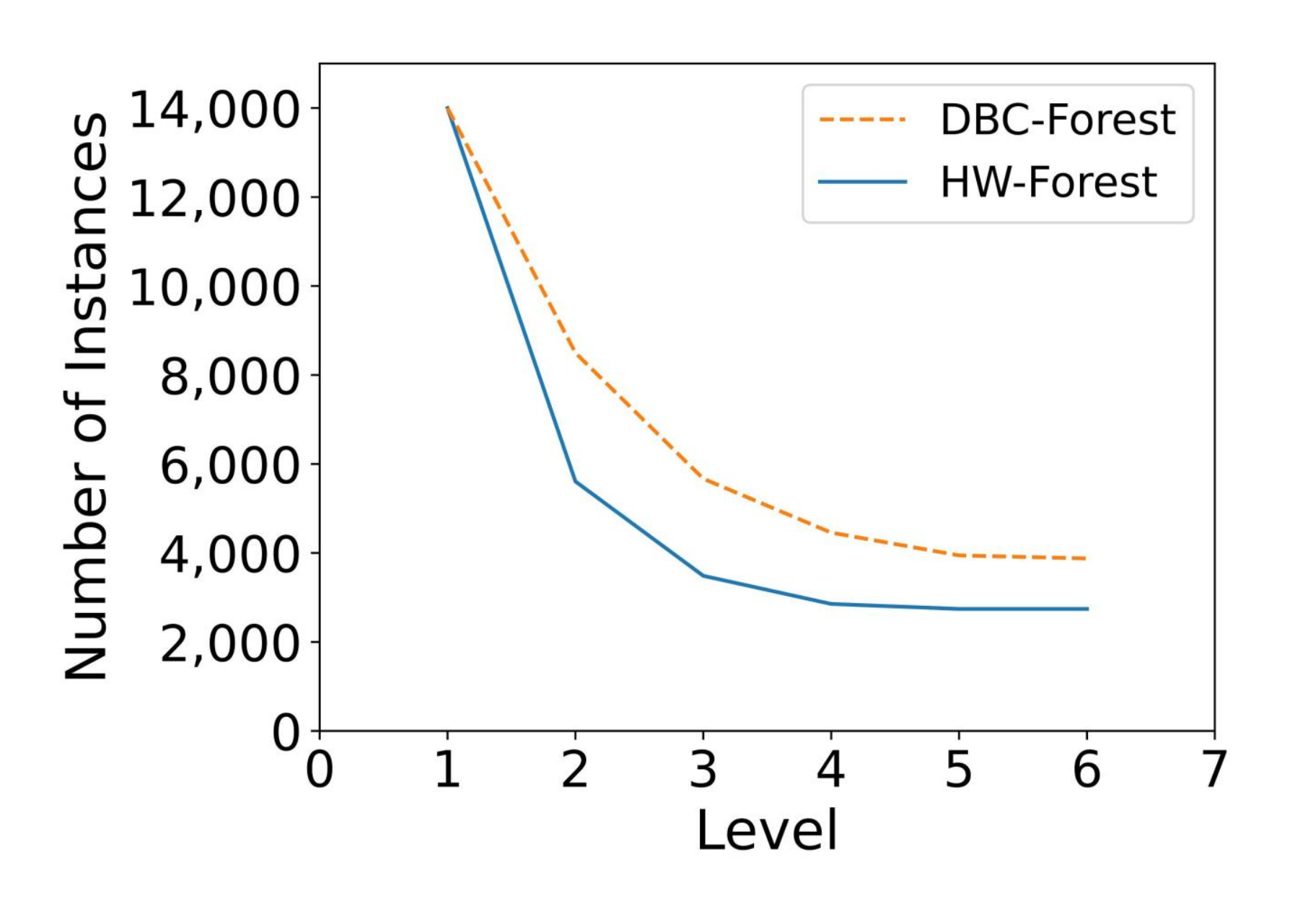}
	
	\caption{Comparison of accuracy and number of instances for the FASHION-MNIST dataset}
	\label{fig:figure11}
\end{figure}

\begin{figure}[H]
	\centering
	\includegraphics[width=0.39\linewidth]{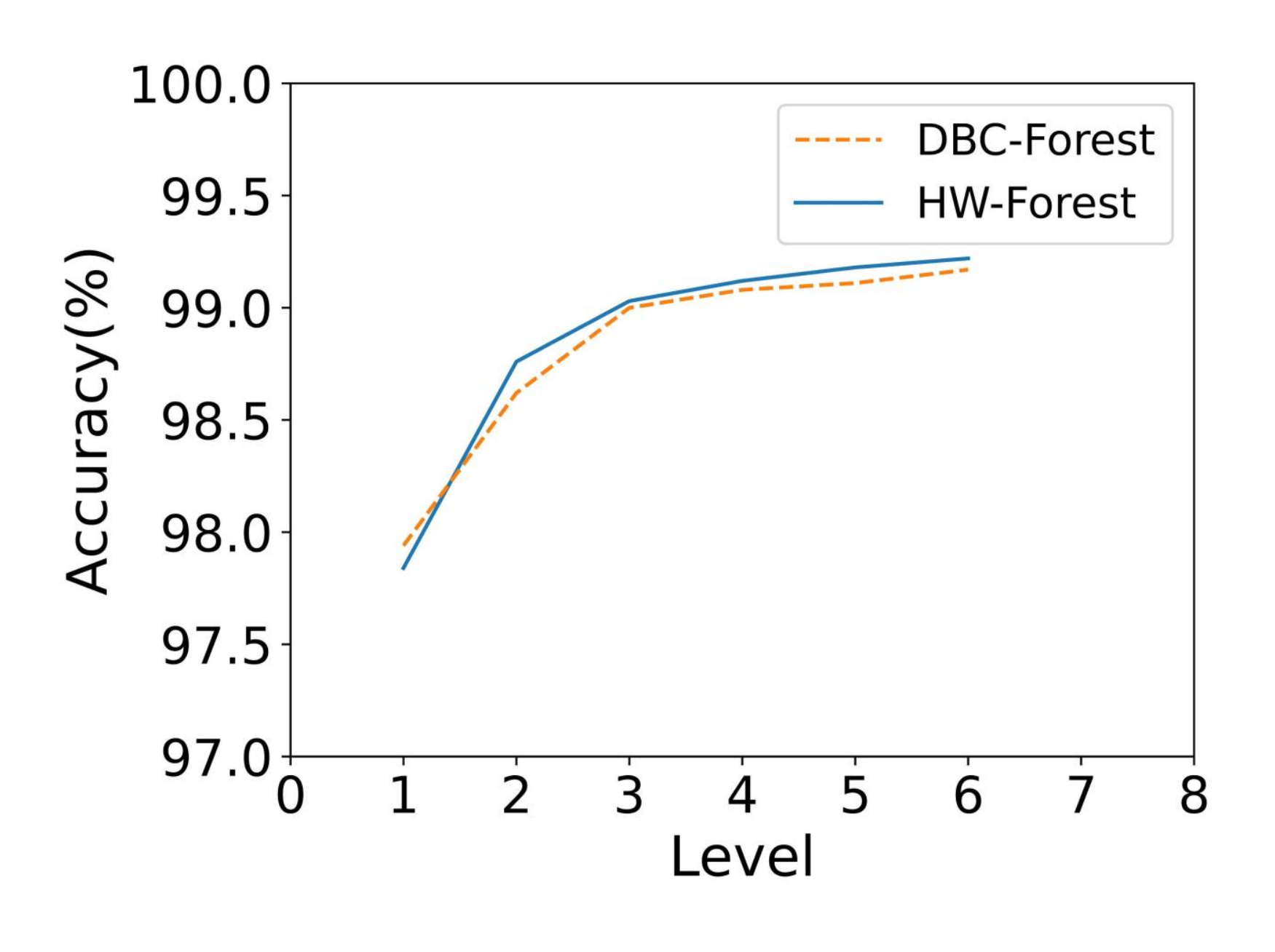}
	\includegraphics[width=0.42\linewidth]{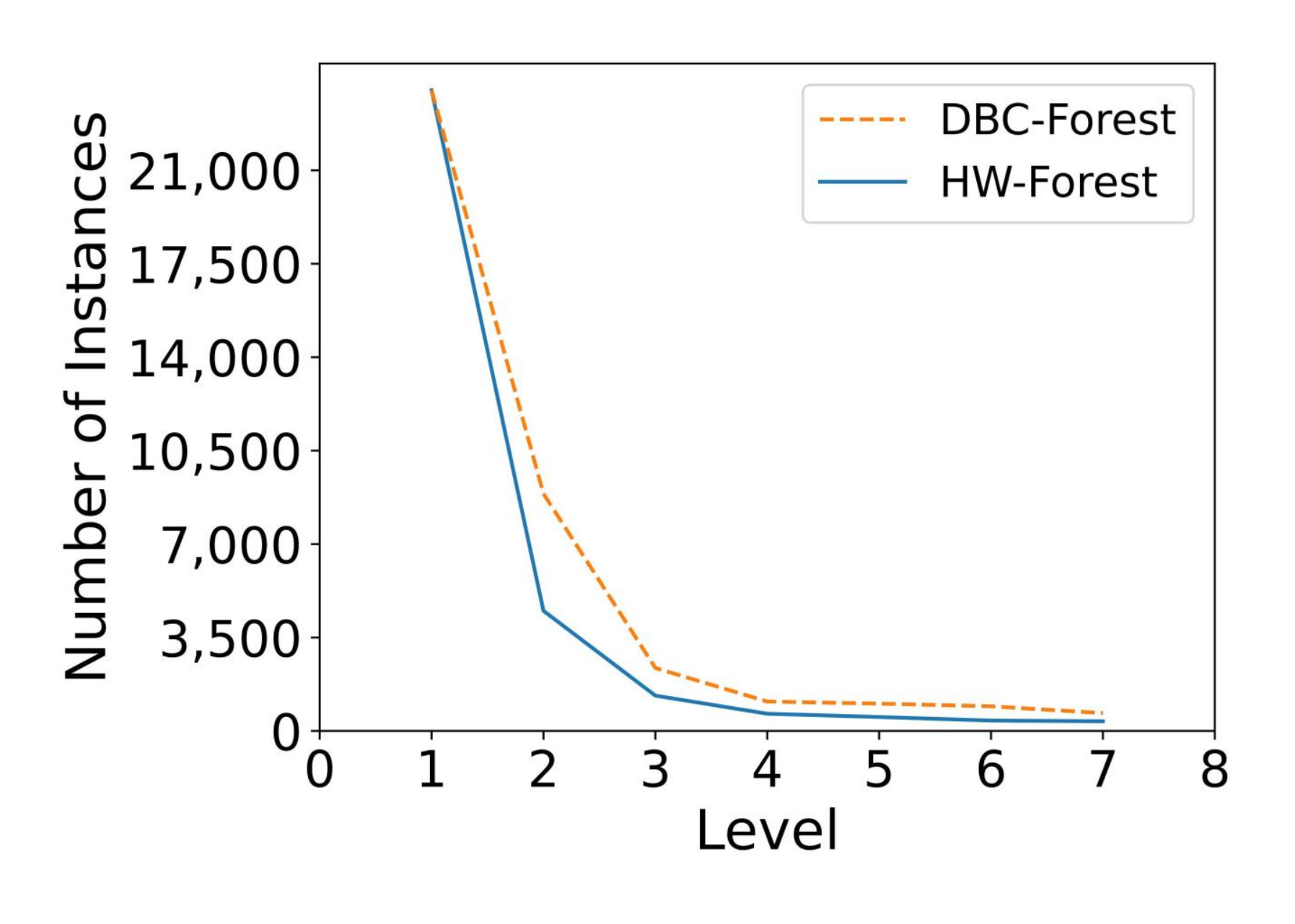}
	
	\caption{Comparison of accuracy and number of instances for the QMNIST dataset}
	\label{fig:figure12}
\end{figure}

\begin{figure}[H]
	\centering
	\includegraphics[width=0.38\linewidth]{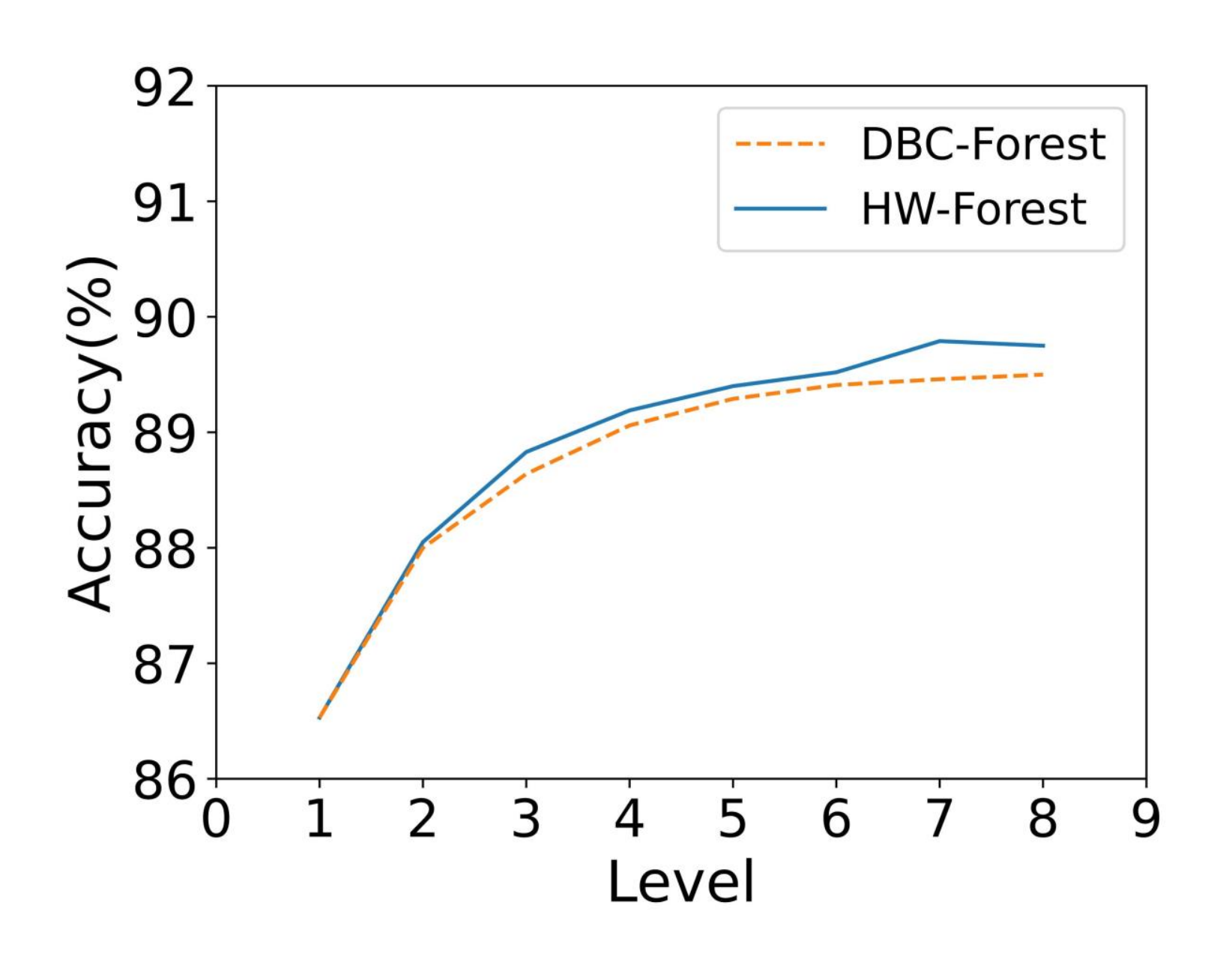}
	\includegraphics[width=0.42\linewidth]{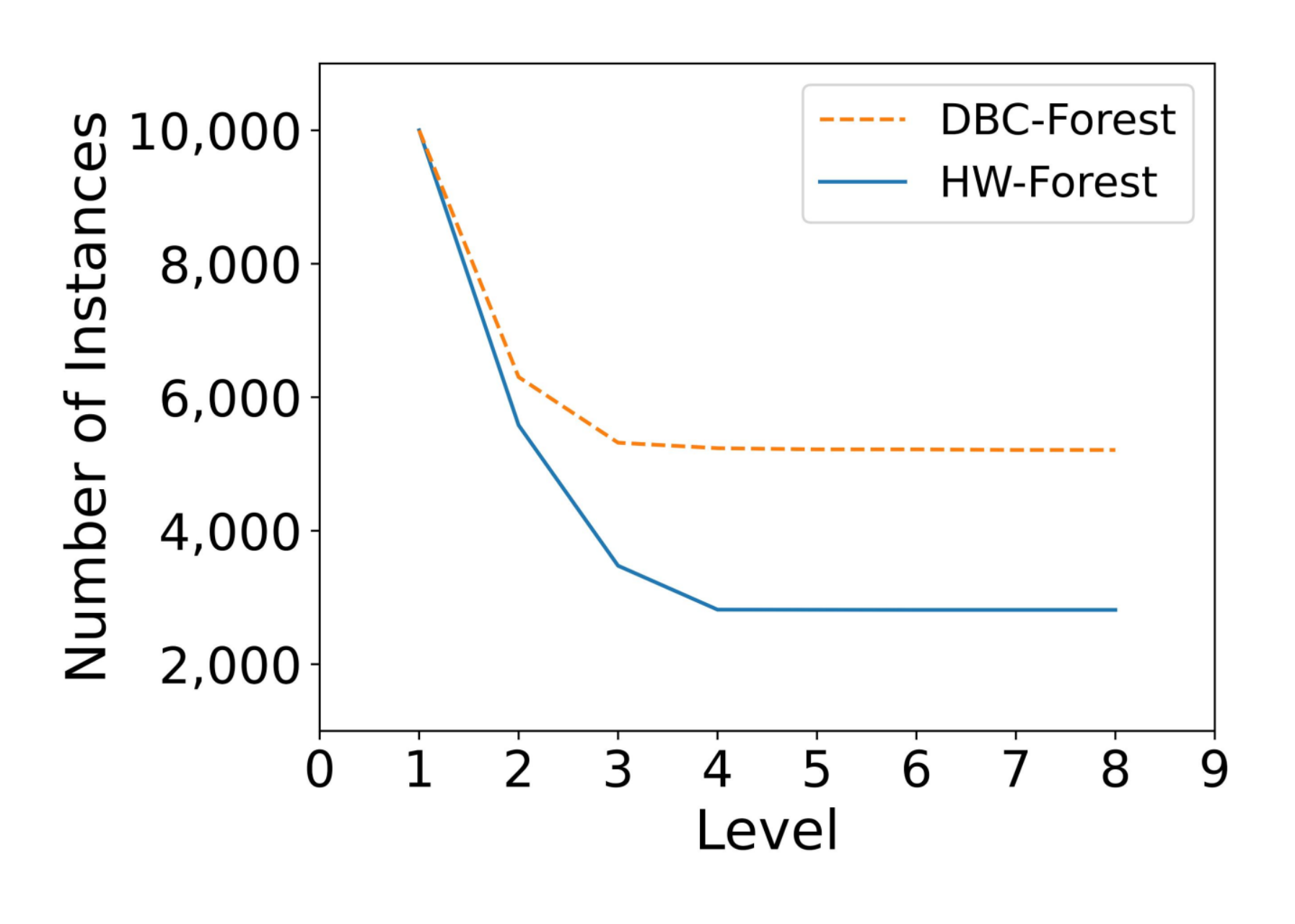}
	
	\caption{Comparison of accuracy and number of instances for the IMDB dataset}
	\label{fig:figure13}
\end{figure}

To validate the effect of window screening, we observed the accuracy and the number of instances at each level. We selected DBC-Forest as a competitive algorithm and used the MNIST, EMNIST, FASHION-MNIST, QMNIST, and IMDB datasets, since our model and competitive algorithm produced more layers for these datasets. Comparisons of the accuracy and numbers of instances for MNIST, EMNIST, FASHION-MNIST, QMNIST, and IMDB are shown in Figures \ref{fig:figure9} to \ref{fig:figure13}, respectively.

These results give rise to the following observations.

1.	HW-Forest achieves higher accuracy than DBC-Forest at each level. For example, it can be seen from Figure \ref{fig:figure13} that for the IMDB dataset, the accuracy of HW-Forest is 0.09\% higher than DBC-Forest at the second level, and 0.19\% higher than DBC-Forest at the last level. The reason for this is that window screening is a self-adaptive mechanism, and produces more reasonable thresholds than the binning screening mechanism without the need for hyperparameter tuning. Hence, the instances screened by HW-Forest have higher accuracy than those screened by DBC-Forest at each level. 

2.	HW-Forest has a lower number of instances than DBC-Forest at each level. For example, on the MNIST dataset, HW-Forest has 1,213 fewer instances than DBC-Forest at the second level, and 146 fewer instances than DBC-Forest at the last level. The reason for this is that since the performance of DBC-Forest depends on the hyperparameters, it is not easy to find high-confidence instances. However, HW-Forest adopts a self-adaptive mechanism to precisely select high-confidence instances.

\subsection{Effect of Window Screening}

\begin{figure}
	\centering
	\includegraphics[width=0.7\linewidth]{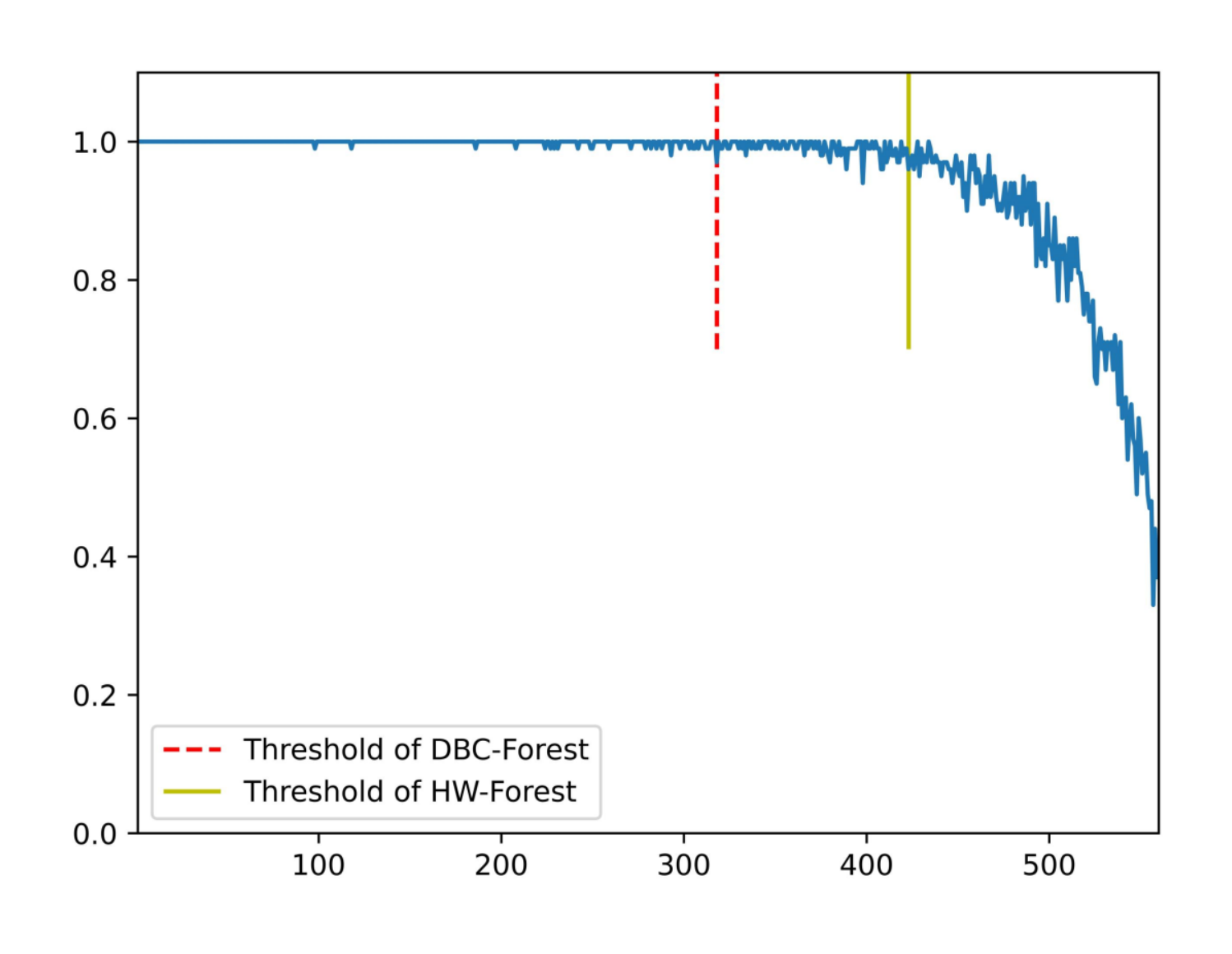}
	\caption{Comparison of thresholds for the MNIST dataset}
	\label{fig:figure14}
\end{figure}

\begin{figure}
	\centering
	\includegraphics[width=0.7\linewidth]{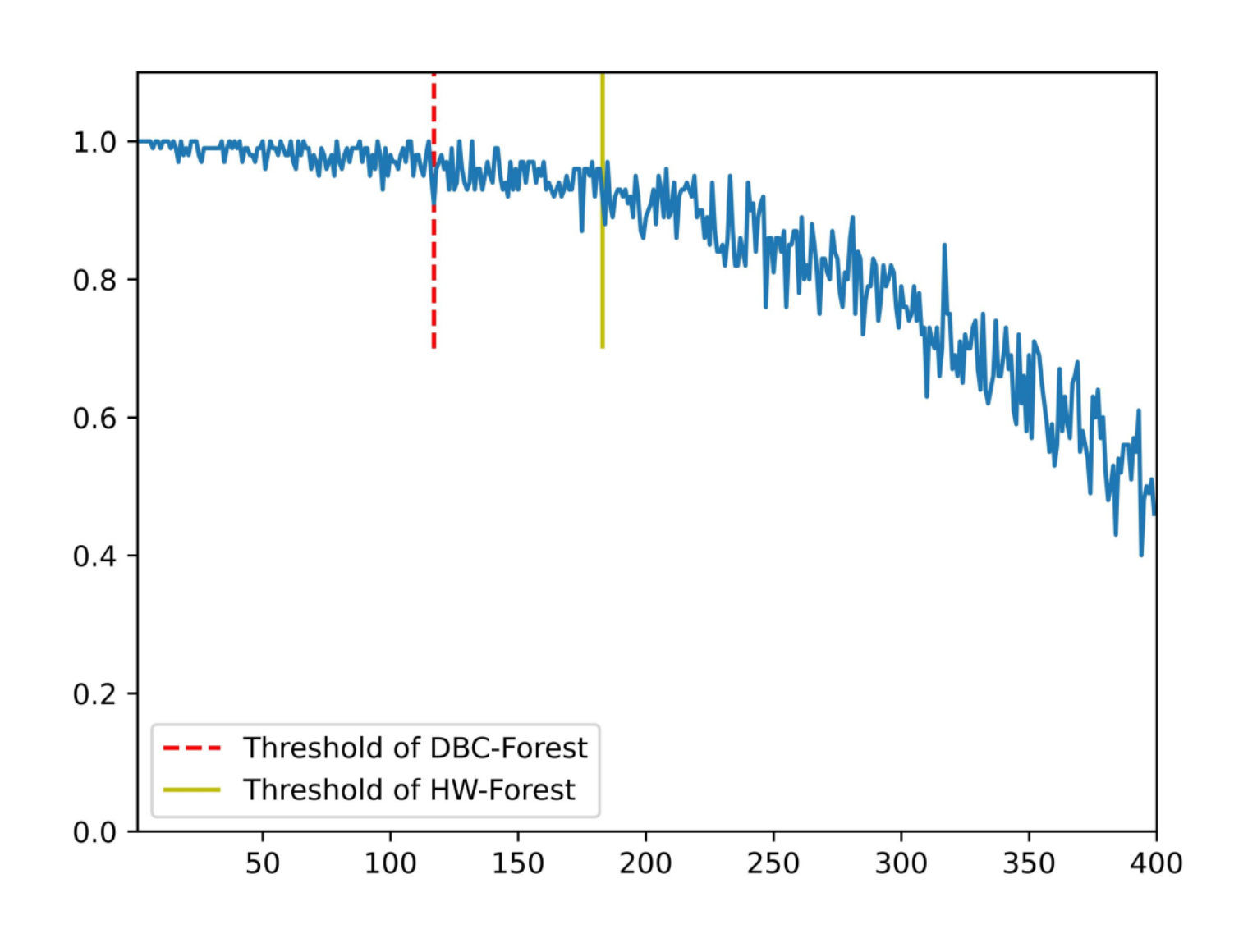}
	\caption{Comparison of thresholds for the IMDB dataset}
	\label{fig:figure15}
\end{figure}

\begin{figure}
	\centering
	\includegraphics[width=0.7\linewidth]{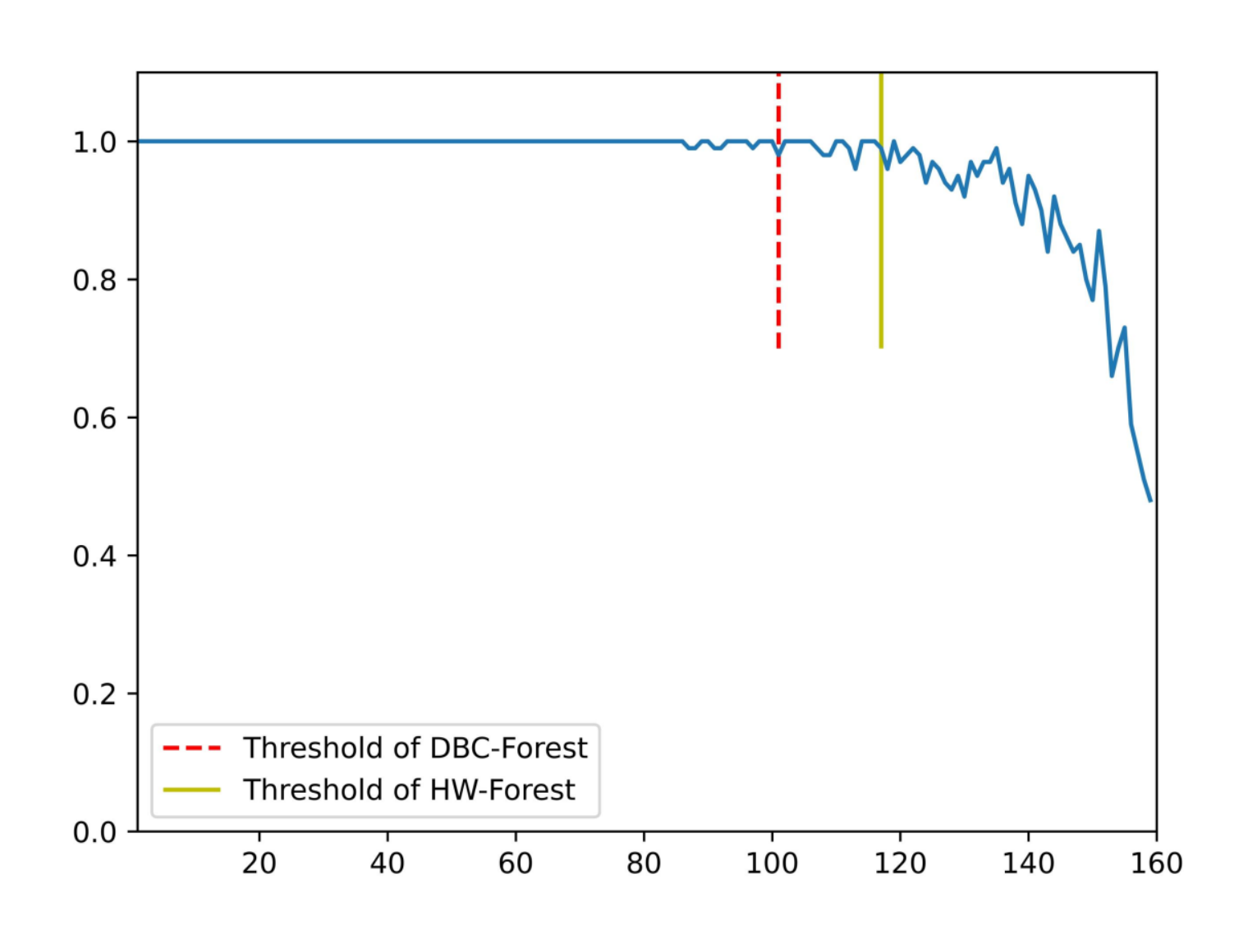}
	\caption{Comparison of thresholds for the LETTER dataset}
	\label{fig:figure16}
\end{figure}

\begin{figure}
	\centering
	\includegraphics[width=0.7\linewidth]{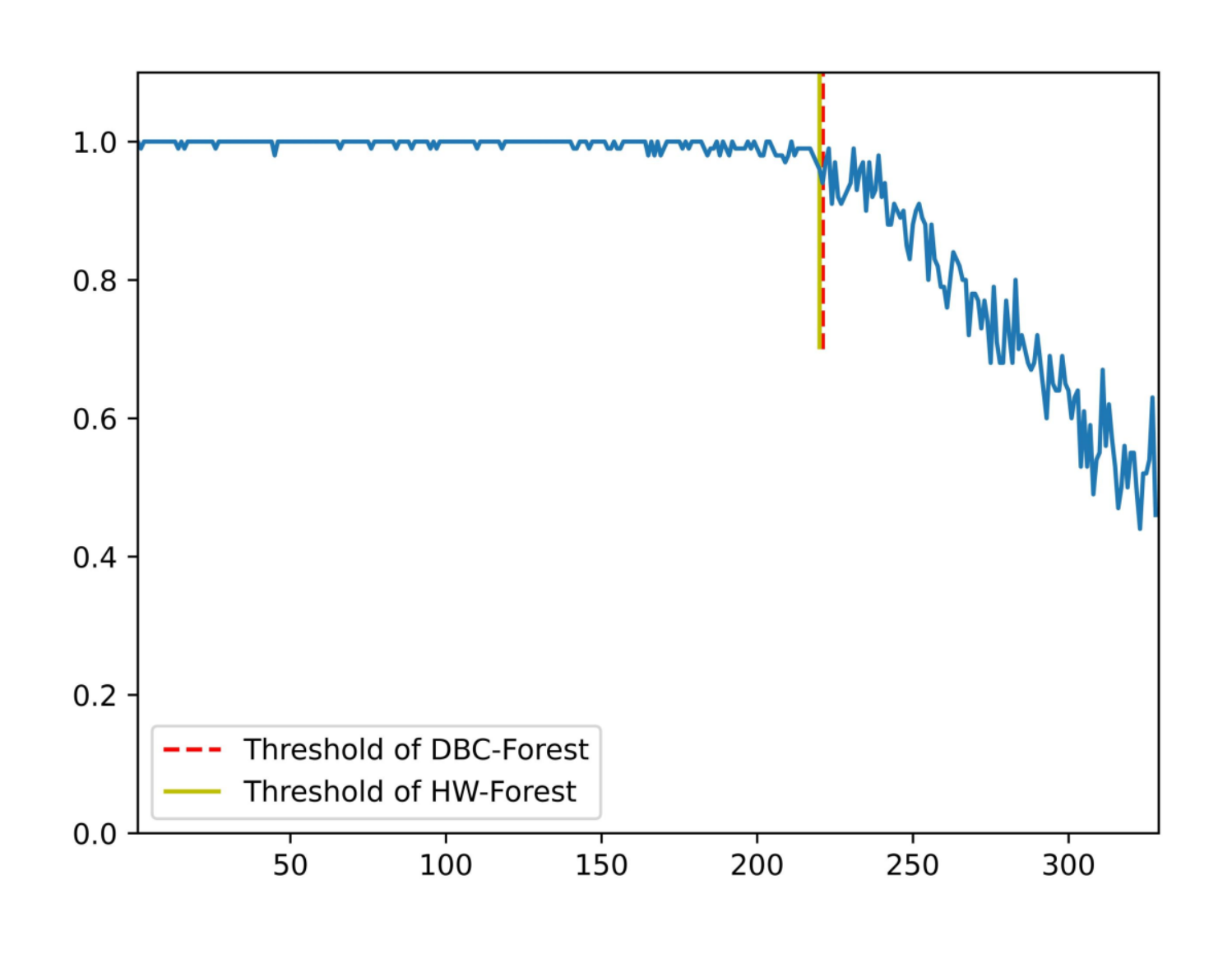}
	\caption{Comparison of thresholds for the BANK dataset}
	\label{fig:figure17}
\end{figure}

In this section, we explore the difference between window screening and binning confidence screening. To demonstrate the performance of our approach on different types of datasets, we used MNIST, IMDB, LETTER, and BANK. Comparisons of the thresholds for the MNIST, IMDB, LETTER, and BANK datasets are shown in Figures \ref{fig:figure14} to \ref{fig:figure17}, respectively.

As shown in Figures \ref{fig:figure14} to \ref{fig:figure17}, HW-Forest gives a more reasonable threshold than DBC-Forest. For example, on the MNIST dataset, DBC-Forest cannot avoid accuracy fluctuations, and some high-confidence instances screened with this threshold have low accuracy. Using HW-Forest, the screened high-confidence instances have significantly higher accuracy. The reason for this is that unlike the dependence of DBC-Forest on hyperparameter tuning, HW-Forest selects instances at each level without the need for hyperparameter tuning. Hence, HW-Forest gives a more reasonable threshold than DBC-Forest.

\subsection{Performance of HW-Forest}

In this section, to validate the accuracy of HW-Forest, we used gcForest, gcForestcs, DBC-Forest, AWDF, H-Forest, and W-Forest as competitive algorithms. Experiments were conducted on nine benchmark datasets, and comparisons of the results for accuracy and time cost are shown in Tables \ref{table10} and \ref{table11}, respectively.

\begin{table*}[]
	\footnotesize 
	\caption{Comparison of accuracy (\%)}
	\centering
	\label{table10}
	\begin{tabular}{@{}
			>{\columncolor[HTML]{FFFFFF}}c 
			>{\columncolor[HTML]{FFFFFF}}c 
			>{\columncolor[HTML]{FFFFFF}}c 
			>{\columncolor[HTML]{FFFFFF}}c 
			>{\columncolor[HTML]{FFFFFF}}c 
			>{\columncolor[HTML]{FFFFFF}}c 
			>{\columncolor[HTML]{FFFFFF}}c 
			>{\columncolor[HTML]{FFFFFF}}c @{}}
		\toprule
		DATASET         & gcforest   & gcForestcs & DBC-Forest & AWDF       & H-Forest   & W-Forest                                                        & HW-Forest                                                       \\ \midrule
		MNIST           & 98.77±0.12 & 98.20±0.09 & 99.03±0.07 & 98.80±0.11 & 98.75±0.12 & 99.05±0.08                                                      & \textbf{99.07±0.09}                                                      \\
		EMNIST          & 86.18±0.24 & 87.11±0.23 & 86.82±0.26 & 86.74±0.18 & 86.13±0.28 & \textbf{87.24±0.18}                                                      & 87.22±0.17                                                      \\
		FASHION-MNIST   & 89.94±0.29 & 89.94±0.29 & 90.57±0.22 & 89.89±0.27 & 90.05±0.29 & 90.77±0.34                                                      & \textbf{90.79±0.24}                                                      \\
		QMNIST          & 98.93±0.06 & 98.40±0.12 & 99.08±0.08 & 98.93±0.08 & 98.92±0.06 & \textbf{99.13±0.05}                                                      & \textbf{99.13±0.05}                                                      \\
		ADULT           & 85.99±0.37 & 86.04±0.35 & 86.11±0.29 & 85.86±0.15 & 85.99±0.37 & \textbf{86.18±0.65}                                                      & \textbf{86.18±0.65}                                                      \\
		BANK  & 91.43±0.23 & 91.58±0.19 & \textbf{91.62±0.16} & 91.45±0.23 & 91.43±0.23 &  91.60±0.22 &  91.60±0.22 
		\\
		YEAST           & 62.06±3.77 & 62.00±3.50 & 62.13±3.70 & 62.53±4.10 & 62.06±3.77 & \textbf{62.67±2.07}                                                      & \textbf{62.67±2.07}                                                      \\
		IMDB            & 88.81±0.10 & 88.88±0.14 & 89.39±0.32 & 88.94±0.23 & 88.81±0.10 & \textbf{89.58±0.26}                                                      & \textbf{89.58±0.26}                                                      \\
		LETTER          & 97.02±0.22 & 96.83±0.27 & 97.07±0.23 & 96.65±0.18 & 97.02±0.22 & \textbf{97.19±0.16}                                                      & \textbf{97.19±0.16}                                                      \\ \bottomrule
	\end{tabular}
\end{table*}

\begin{table*}[]
	\footnotesize 
	\centering
	\caption{Comparison of time cost (s)}
	\label{table11}
	\begin{tabular}{c c c c c c c c }
		\toprule
		DATASET         & gcforest       & gcForestcs     & DBC-Forest     & AWDF           & H-Forest      & W-Forest      & HW-Forest     \\ \midrule
		MNIST           & 1972.27±6.39   & 1636.00±13.47  & 1636.10±17.79  & 1697.79±18.36  & 1464.86±11.54 & 1648.92±20.56 & \textbf{1135.42±9.32}  \\
		EMNIST          & 8774.77±108.08 & \textbf{5945.37±159.68} & 7926.23±112.34 & 7385.44±103.03 & 6920.12±51.39 & 6856.64±80.16 & 6408.14±65.33 \\
		FASHION-MNIST   & 2411.10±10.28  & 1943.09±6.50   & 2056.95±13.38  & 1821.65±15.30  & 2272.63±25.18 & 1990.47±13.91 & \textbf{1904.17±22.82} \\
		QMNIST          & 2332.79±31.80  & 1991.11±17.71  & 1979.00±22.14  & 1988.55±66.69  & 1790.10±17.71 & 1965.98±34.68 & \textbf{1473.09±12.00} \\
		ADULT           & 10.63±0.18     & \textbf{9.88±0.08}      & 10.18±0.23     & 10.21±0.12     & 10.63±0.18    & 10.32±0.18    & 10.15±0.08    \\
		BANK  & 12.68±0.13     & \textbf{9.83±0.26}      & 10.84±0.07     & 11.64±0.06     & 12.77±0.16    & 11.86±0.11    & 12.056±0.12   \\
		YEAST           & 8.52±0.04      & \textbf{8.47±0.07}      & 8.56±0.09      & 10.65±0.23     & 8.52±0.04     & 8.69±0.3      & 8.69±0.33     \\
		IMDB            & 1,257.24±6.28  & \textbf{475.65±10.53}   & 753.52±12.18   & 1,109.72±16.22 & 1,257.24±6.28 & 551.96±8.41   & 551.96±8.41   \\
		LETTER          & 11.89±0.18     & \textbf{9.02±0.10}      & 9.02±0.10      & 10.68±0.12     & 11.89±0.18    & 9.63±0.14     & 9.63±0.14     \\ \bottomrule
	\end{tabular}
\end{table*}

From these results, we can make the following observations.

1.	As shown in Table \ref{table10}, HW-Forest achieves the highest accuracy on most datasets. For example, on the QMNIST dataset, HW-Forest achieves the highest accuracy of 99.07\%. The same phenomenon can be found for the other datasets. The reason for this is that HW-Forest uses window screening to improve the accuracy of DBC-Forest, and this produces more reasonable thresholds than the binning screening mechanism. Hence, HW-Forest achieves higher accuracy than DBC-Forest, and also achieves higher accuracy than other models. As shown in Table \ref{table11}, for high-dimensional datasets, HW-Forest runs faster than the competitive models in most cases; for low-dimensional datasets, HW-Forest runs slower than gcForestcs, but faster than the other competitive models. For example, on the MNIST dataset, the time cost for HW-gcForest is 1135.42 s, which is faster than all other models. On the ADULT dataset, the time cost of HW-gcForest is 10.15 s, which is slower than gcForestcs but faster than the other competitive models. The reason for this is that for high-dimensional datasets, deep forest models take a long time to find the classification models. If we can reduce the number of feature vectors, the running time can be significantly reduced. HW-Forest employs hashing screening to effectively eliminate redundant feature vectors, and therefore runs faster than the competitive models for high-dimensional datasets. However, for low-dimensional datasets, the deep forest models take only a short time to find the classification models, and these datasets do not need to be processed by multi-grained scanning. Hence, the running time of precisely selecting instances at each level cannot be neglected, and HW-Forest runs slower than gcForestcs. 

2.	HW-Forest and W-Forest achieve the same accuracy on most datasets. For example, from Table \ref{table10}, we can see that for the QMNIST dataset, HW-Forest and W-Forest achieve the same accuracy of 99.13\%. The same effect can be seen for the ADULT, BANK, YEAST, IMDB, and LETTER datasets. For the MNIST, FASHION-MNIST, and EMNIST datasets, the results for the accuracy are close. The reason is for this is that low-dimensional datasets do not need to be processed by multi-grained scanning. The results of HW-Forest are therefore the same as those of W-Forest, since W-Forest is simply the HW-Forest model without the hashing screening mechanism. For high-dimensional datasets, as demonstrated in our analysis of the effect of hashing screening in Section 5.3, the hashing screening mechanism does not influence the accuracy of the model. Hence, the accuracies of HW-Forest and W-Forest are almost the same.

To further evaluate the performance of DBC-Forest, we applied Friedman and Nemenyi tests \cite{wu2021tcy} for statistical hypothesis test.

1. Friedman test: This test ranks the models according to their accuracy, where the highest is given a score of one, the second highest has a score of two, and so on. The mean values for the gcForest, gcForestcs, DBC-Forest, AWDF, HW-Forest, H-Forest, and W-Forest models were 5.89, 5.33, 3, 5.11, 1.61, 5.44 and 1.61, respectively. We then calculated the Friedman test statistic according to  $(N-1)\times \tau_{\chi^2}/(N\times (k-1)-\tau_{\chi^2})$, where $\tau_{\chi^2}=12\times N\times (\sum_{i=1}^{k}r_i^2-k\times (k+1)^2/4)/k\times (k+1)$, and $N$ and $k$ are the number of datasets and models, respectively. The Friedman statistic was 24.371, which was larger than $T_{0.05,4}$ = 2.132. Hence, we reject the null hypothesis. 

\begin{figure}
	\centering
	\includegraphics[width=0.7\linewidth]{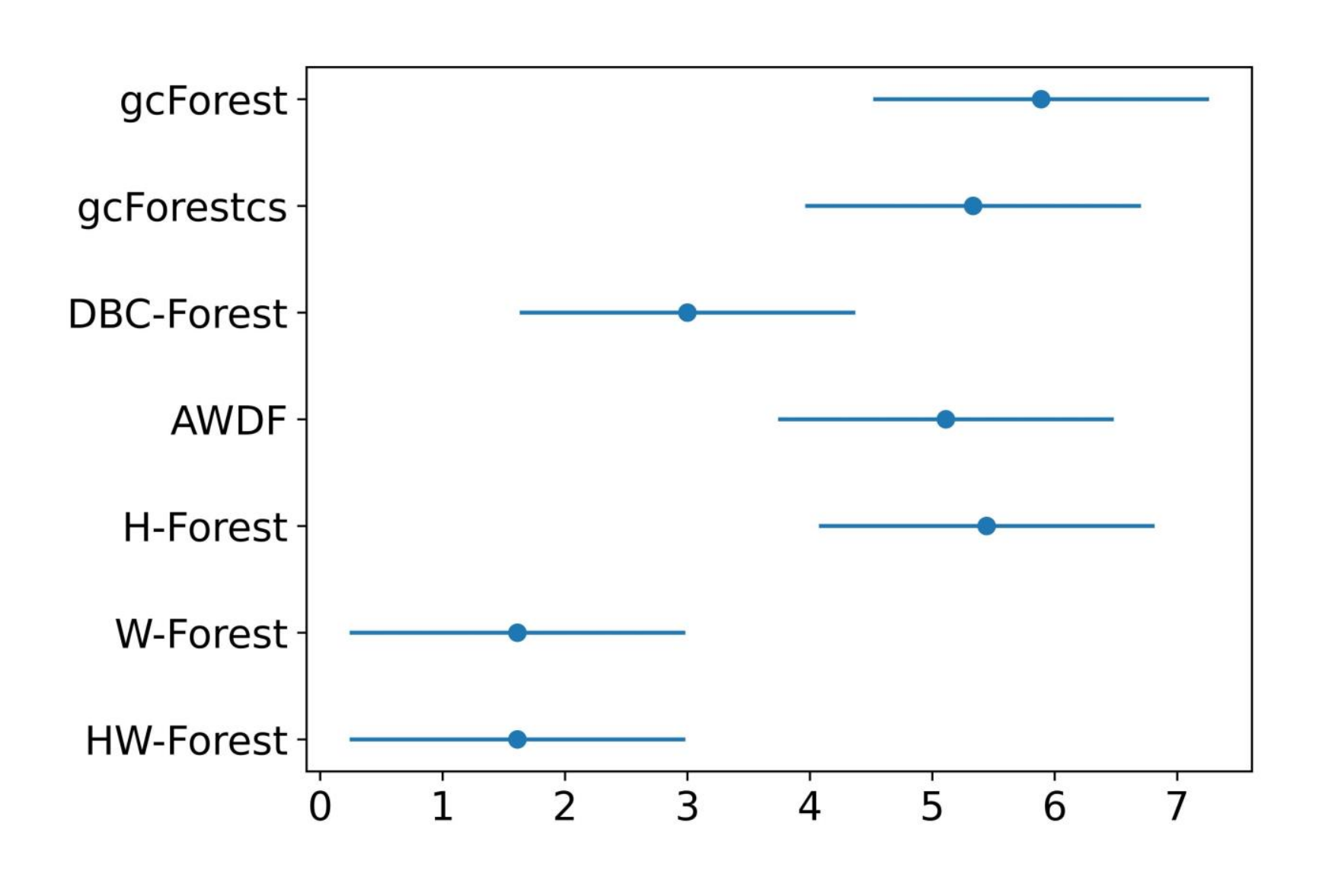}
	\caption{Results of a Nemenyi test, where the dots show the mean value for each model, and the horizontal bar across each dot denotes the critical difference}
	\label{fig:figure18}
\end{figure}

2. Nemenyi test: To further compare the models, we adopted the Nemenyi test to calculate the critical difference value. The result was 2.742, according to the formula $q_{0.1,7}\sqrt{k(k+1)/6N}$, where  $q_{0.1,7}$ = 2.693. As shown in Figure \ref{fig:figure18}, the accuracy of HW-Forest is the same as for W-Forest, which indicates that HW-Forest is statistically consistent with W-Forest. More importantly, the accuracy of HW-Forest is significantly higher than the other models except W-Forest, meaning that HW-Forest statistically outperformed the other competitive models. 

\section{Conclusion}
To improve the performance and efficiency of the deep forest model, we propose a novel deep forest model called HW-Forest in which two mechanisms are applied to improve the performance: hashing screening and window screening. Hashing screening employs the feature distance as the criterion for selecting feature vectors, and this is calculated using a perceptual hashing algorithm. The number of redundant feature vectors can therefore be reduced by hashing screening. We also propose a window screening mechanism which produces the threshold by using a self-adaptive process for the length of the window. This method can effectively find the threshold without using an unsuitable hyperparameter for the size of the bins. Our experimental results demonstrate that hashing screening effectively reduces the time cost of multi-grained scanning, and the window screening mechanism outperforms the binning confidence screening mechanism. More importantly, our experimental results indicate that HW-Forest gives better performance than other competitive models, since it is statistically significantly better than these models. 

\section*{Acknowledgement}
This work was supported by National Natural Science Foundation of China (61976240, 52077056, 91746209),  National Key Research and Development Program of China (2016YFB1000901) , Natural Science Foundation of Hebei Province, China (Nos. F2020202013, E2020202033), and Graduate Student Innovation Program of Hebei Province (CXZZSS2021026).

\end{document}